\pdfoutput=1
\documentclass[10pt,twocolumn,letterpaper]{article}

\usepackage[pagenumbers]{cvpr} 
\usepackage{multirow}
\usepackage{algorithm}
\usepackage{algpseudocode}
\usepackage[accsupp]{axessibility}

\definecolor{cvprblue}{rgb}{0.21,0.49,0.74}
\usepackage[pagebackref,breaklinks,colorlinks,allcolors=cvprblue]{hyperref}

\def\paperID{ } 
\def\confName{CVPR}
\def\confYear{2026}

\title{Local Motion Matters: A Deconstruct–Recompose Paradigm for Reinforcement Learning Pre-training from Videos}

\author{
Jinwen Wang$^{1,2}$ \quad
Youfang Lin$^{1,2}$ \quad
Xiaobo Hu$^{1,2}$ \quad
Shuo Wang$^{1,2}$\thanks{Corresponding author.} \quad
Kai Lv$^{1,2}$\\[0.3em]
$^{1}$Beijing Jiaotong University, China\\
$^{2}$Beijing Key Laboratory of Traffic Data Mining and Embodied Intelligence, China\\[0.2em]
{\tt\small \{jw.wang, yflin, xiaobohu, shuo.wang, lvkai\}@bjtu.edu.cn}
}

\begin{document}
\maketitle
\begin{abstract}
Pre-training on large-scale videos to improve reinforcement learning efficiency is promising yet remains challenging. 
Existing methods typically treat the agent as an indivisible entity, modeling motion patterns globally.
Such global modeling is tightly coupled with the morphology, hindering transfer across domains. 
In contrast, despite the vast disparity in global motions, the local components exhibit similar motion patterns across different agents.
Building on this insight, we propose a novel Deconstruct–Recompose Paradigm (DRP) for learning transferable local motion representations. 
Specifically, in the Deconstruct phase, we identify multiple local points and track their frame-wise motions, defining each as an Atomic Action. 
We introduce a Dual-Attention Encoder (DAE) to learn local motion representations from these Atomic Actions, capturing their spatiotemporal relationships.
In the Recompose phase, we compose local motion representations with a learnable Motion Aggregation Token `[MAT]' via latent dynamics model learning.
Additionally, an adapter bridges local motion and downstream action-specific dynamics to accelerate policy learning.
Extensive experiments demonstrate that our method effectively transfers to diverse robotic control and manipulation tasks, significantly improving sample efficiency and performance.
\end{abstract}    
\section{Introduction}
\label{sec:intro}

\begin{figure}
    \centering
 \includegraphics[width=0.48\textwidth]{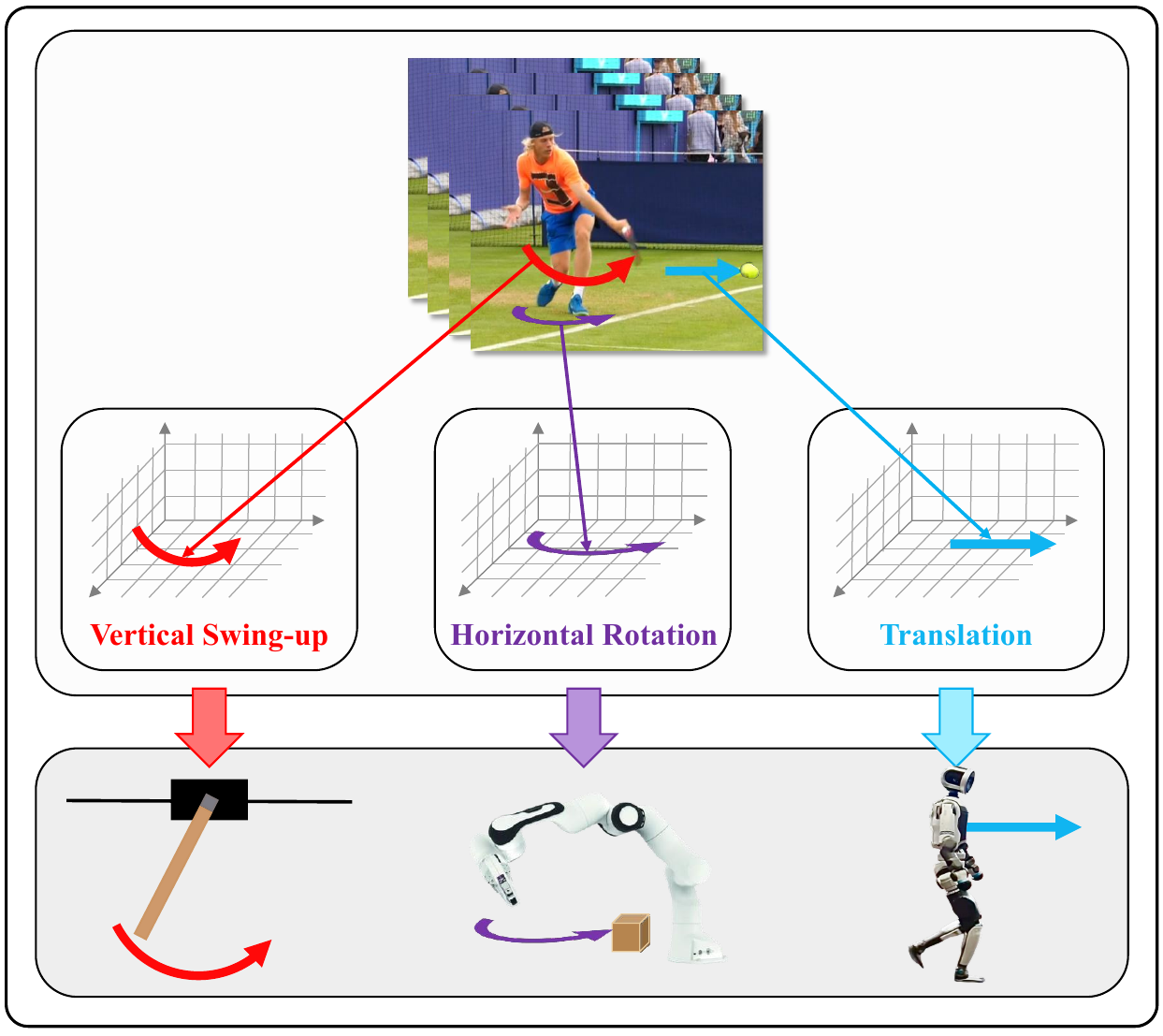}
    \caption{\textbf{Illustration of local
 motion patterns.} 
 The global motion of a tennis player’s forehand swing (top) is composed of several local motions (middle), such as the arm’s vertical swing-up, the body’s horizontal rotation, and the ball’s translation, which more easily capture similarities across tasks (bottom).}
    \label{fig:intro}
\end{figure}

Deep Reinforcement Learning (RL) has achieved remarkable success in various fields~\cite{DBLP:journals/jmlr/LevineFDA16, DBLP:conf/icml/LaskinSA20, DBLP:conf/nips/ChenWWFJLMDZY22, DBLP:conf/nips/HafnerLFA22, DBLP:journals/tmm/WangWHLL23, DBLP:journals/corr/abs-2309-13041, 
DBLP:conf/ijcai/Wang0WHLL24,
DBLP:conf/eccv/PengLLSGSF24, DBLP:journals/tase/CuiXZXST25, DBLP:conf/cvpr/ZhangLGL0XM025}. 
However, RL typically trains from scratch, requiring extensive online interactions to acquire effective policies. 
To mitigate this fundamental challenge,
recent works have explored leveraging large-scale, unlabeled video datasets for pre-training~\cite{DBLP:conf/corl/NairRKF022,DBLP:conf/icml/SeoLJA22, DBLP:conf/nips/BakerAZHTEHSC22, DBLP:conf/corl/RadosavovicXJAM22, DBLP:conf/iclr/MaSJBK023, DBLP:conf/icml/GhoshBL23,DBLP:conf/nips/0001MDL23,DBLP:conf/mm/WangLHYH0025}. 
This pre-training paradigm aims to provide downstream agents with rich \textit{knowledge priors}, thereby significantly improving sample efficiency and performance.


Current unsupervised pre-training approaches for RL primarily fall into two categories. 
The first one is to pre-train a video prediction model, aiming to help downstream agents understand environment dynamics~\cite{DBLP:conf/icml/SeoLJA22, DBLP:conf/nips/0001MDL23,wang2025disentangled}. 
The second approach infers latent actions as factors of environment dynamics from inter-frame relations, facilitating downstream policy learning~\cite{DBLP:journals/corr/abs-1901-03162, DBLP:conf/iclr/YeZAG23, DBLP:conf/iclr/SchmidtJ24, DBLP:conf/eccv/LuoZL24, DBLP:conf/eccv/ZhangKSC24}. 
These approaches model motion patterns at a \textit{global} level, treating the agent as an indivisible entity.
Therefore, the learned motion patterns are tightly coupled with the morphology, limiting cross-domain transfer.



In fact, while the global motions of different agents vary significantly, the local motion patterns often exhibit strong similarities.
As illustrated in Figure~\ref{fig:intro}, the global motion of a tennis player’s forehand swing differs markedly from that of downstream tasks.
However, this complex global motion is composed of local components: the arm’s Vertical Swing-up, the body’s Horizontal Rotation, and the ball’s Translation.
These local components are more similar across different agents.
This contrast between \textit{global disparity} and \textit{local similarity} suggests a core insight: \textit{``To achieve robust cross-domain transfer, can we employ a local modeling paradigm instead of global modeling''?}

To this end, we propose a novel \textit{Deconstruct-Recompose Paradigm} (DRP) that realizes this local modeling paradigm.
In the Deconstruct phase, we first deconstruct morphology-coupled global motion into a set of morphology-agnostic local motion components, termed \textit{Atomic Actions}.
Then, a Dual-Attention Encoder (DAE) is introduced to learn transferable local motion representations based on these Atomic Actions.
In the Recompose phase, we employ a learnable Motion Aggregation Token `[MAT]' to compose local motion representations, and enrich them with dynamic semantics via latent dynamics model learning.
In practice, the \textit{Deconstruct} phase involves two key sub-processes.
First, we extract \textit{Atomic Actions} by tracking persistent keypoints and cropping surrounding local optical flow patches. 
Second, to learn local motion representations, the DAE incorporates Intra-Frame and Inter-Frame attention for capturing the spatiotemporal relationships among these Atomic Actions.
The DAE is trained using a Masked Autoencoder (MAE) reconstruction objective.
In the \textit{Recompose} phase, a learnable `[MAT]' token composes local motion representations to align with the specific task. 
Specifically, we apply a latent dynamics model to capture temporal relationships, enriching the composed representations with dynamic semantics. 
Additionally, an adapter maps these representations to the downstream specific action space, accelerating policy learning.

Our main contributions are summarized as follows:
\begin{itemize}



    \item We introduce a novel Deconstruct-Recompose Paradigm (DRP) that focuses on modeling local motion patterns rather than global motions, enabling effective cross-domain transfer.




    \item We design a novel Dual Attention Encoder (DAE) that incorporates Intra-Frame and Inter-Frame attention to learn local motion representations, enabling capturing spatiotemporal relationships.

    

    \item Extensive experiments demonstrate that our pre-training method based on local motion representations significantly improves sample efficiency and performance across various robotic control and manipulation tasks.

\end{itemize}

\section{Related Work}
\label{sec:related_work}

\subsection{Model-Based Reinforcement Learning}
Model-based Reinforcement Learning (MBRL) improves sample efficiency by constructing a world model of the environment to generate imagined trajectories~\cite{DBLP:conf/iclr/HafnerLB020, DBLP:conf/iclr/HafnerL0B21, hafner2025dreamerv3}.

PlaNet introduces a Recurrent State Space Model (RSSM) to learn environment dynamics from images and selects actions through planning in latent spaces.
The Dreamer series further advances this research direction.
Dreamer~\cite{DBLP:conf/iclr/HafnerLB020} proposes a latent dynamics model based on a Variational AutoEncoder (VAE)~\cite{DBLP:journals/corr/KingmaW13}, which encodes observations and actions into compact latent states and effectively learns policies from imagined latent trajectories.
DreamerV2~\cite{DBLP:conf/iclr/HafnerL0B21} introduces a discretized world model, achieving human-level performance on the Atari benchmark for the first time. 
DreamerV3~\cite{hafner2025dreamerv3} further demonstrates remarkable generalization, surpassing specialized methods across more than 150 diverse tasks using a single set of hyperparameters.
Additionally, approaches such as TD-MPC~\cite{DBLP:conf/icml/HansenSW22} advance MBRL by combining the strengths of model predictive control and efficient temporal-difference learning.

Despite these advances, traditional MBRL approaches typically learn from scratch and require extensive online interactions to build accurate models.
This limitation motivates the exploration of pre-training on large-scale, unlabeled video data.

\subsection{RL Pre-training via Video Prediction}
Recent studies~\cite{DBLP:conf/corl/RadosavovicXJAM22, DBLP:conf/icml/ParisiRP022, DBLP:conf/icml/GhoshBL23,DBLP:conf/corl/NairRKF022} demonstrate that unsupervised Reinforcement Learning (RL) pre-training from videos significantly improves sample efficiency and asymptotic performance across various downstream decision-making tasks.
One line of research focuses on pre-training a video prediction model.

APV~\cite{DBLP:conf/icml/SeoLJA22} proposes an action-free world model pre-trained with videos from different domains: RLBench~\cite{DBLP:journals/ral/JamesMAD20}. 
Building on this, IPV~\cite{DBLP:conf/nips/0001MDL23} extends this paradigm to handle complex ``in-the-wild'' videos by introducing a contextualized world model that explicitly disentangles static context from temporal dynamics. 
Other works, such as DisWM~\cite{wang2025disentangled} pre-train task-relevant representations from distracting videos and transfer them to downstream RL via latent distillation.

\begin{figure*}
    \centering
    \includegraphics[width=0.96\textwidth]{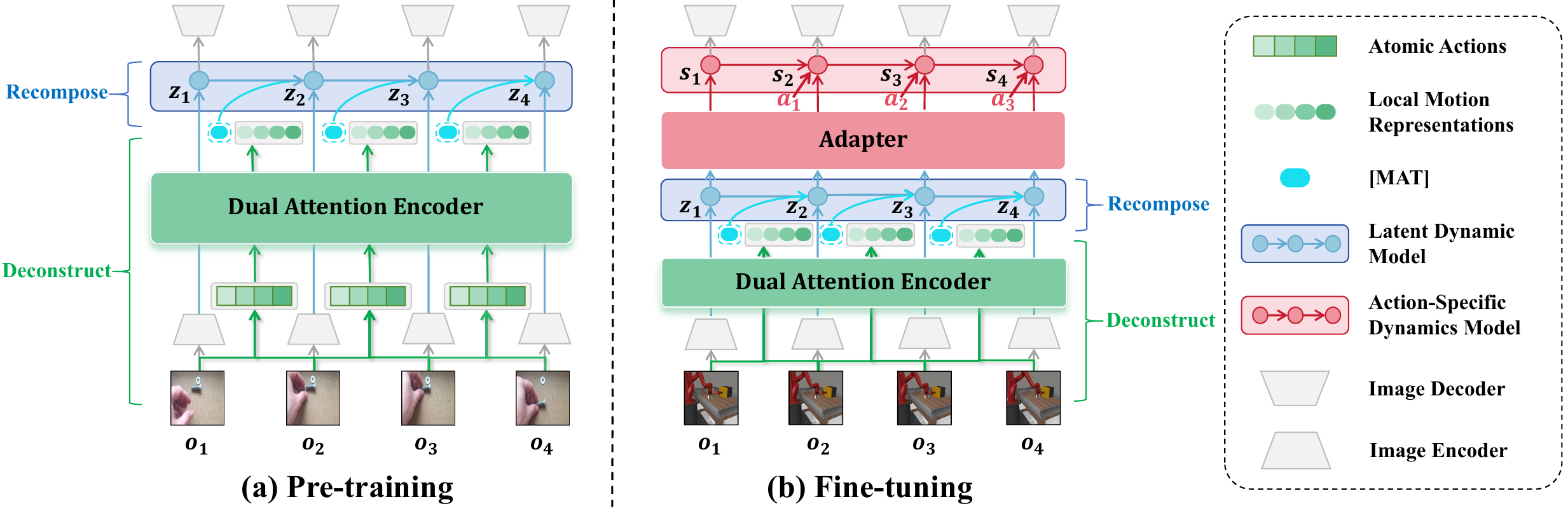}
    \caption{\textbf{Framework Overview.}
    \textbf{(a) Pre-training}: We deconstruct global motion into local Atomic Actions and learn transferable local motion representations, followed by recomposing local motion representations via a latent dynamics model learning.
    \textbf{(b) Fine-tuning}: The DRP is progressively adapted, and an adapter maps the recomposed local motion representations to the agent-specific action space.}
    \label{fig:architecture}
\end{figure*}

\subsection{RL Pre-training via Latent Action Modeling}
Another line of research focuses on inferring latent actions from inter-frame relations as factors of the dynamic model.

FICC~\cite{DBLP:conf/iclr/YeZAG23} introduces a forward-inverse cycle consistency objective, which jointly infers latent actions and trains the dynamics model. 
AVDC~\cite{DBLP:conf/iclr/KoMDST24} computes optical flow between video frames and combines it with depth information to infer closed-form actions to be executed in the environment.
PVDR~\cite{DBLP:conf/eccv/LuoZL24} employs a Conditional Variational Autoencoder (CVAE)~\cite{DBLP:conf/nips/SohnLY15} to learn visual dynamics representations from videos, based on which it trains a dynamic model. 
PreLAR~\cite{DBLP:conf/eccv/ZhangKSC24} proposes learning action representations by encoding two consecutive observations through an inverse dynamics model, serving as the action factors of the dynamic model.

Whether these methods learn a video prediction model or a latent action–based dynamics model, they all perform motion modeling at a global level.
Consequently, the learned motion patterns become tightly coupled with the agent’s morphology, thereby hindering effective cross-domain transfer.


\section{Preliminaries}
\label{sec:preliminaries}

\subsection{Problem Formulation}
We formulate the visual control task as a Partially Observable Markov Decision Process (POMDP), defined by the tuple $\mathcal{M} = \langle \mathcal{O}, \mathcal{A}, p, r, \gamma \rangle$.
Here, $\mathcal{O}$ is the high-dimensional observation space, $\mathcal{A}$ is the action space, $p(o_{t+1}|o_t, a_t)$ is the environment transition dynamics, $r(o_t, a_t)$ is the reward function, and $\gamma \in [0, 1)$ is the discount factor. The objective of RL is to learn a policy $\pi(a_t | o_t)$ that maximizes the expected cumulative discounted reward $\mathbb{E}[\sum_{t=1}^{T} \gamma^t r_t]$.


\subsection{Latent Dynamics Models}
Dreamer~\cite{DBLP:conf/iclr/HafnerLB020} proposes the latent dynamics model, a world model that effectively learns from high-dimensional images, consisting of several key components:

\begin{itemize}
    \item \textbf{Representation Model:} $z_t \sim q_\phi(z_t | z_{t-1}, a_{t-1}, o_t)$. This model, also known as the posterior, infers the current latent state $z_t$ from the current observation $o_t$ and the previous state and action.
    \item \textbf{Transition Model:} $\hat{z}_t \sim p_\phi(\hat{z}_t | z_{t-1}, a_{t-1})$. This model, also known as the prior, predicts the next latent state purely from the previous state and action, enabling imagination of future trajectories.
    \item \textbf{Image Decoder:} $\hat{o}_t \sim p_\phi(\hat{o}_t | z_t)$. Reconstructs the observation from the latent state, ensuring that $z_t$ captures sufficient visual information.
    \item \textbf{Reward Predictor:} $\hat{r}_t \sim p_\phi(\hat{r}_t | z_t)$. Predicts the reward from the latent state.
\end{itemize}




\subsection{Optical Flow as Motion Representation}
Optical flow provides a dense, pixel-wise description of motion between consecutive video frames.
For two consecutive image frames $o_t$ and $o_{t+1}$, a dense optical flow field $F_{t}$ is a 2D vector field where each vector $(u, v)$ at pixel location $(x, y)$ represents the displacement of that pixel from frame $t$ to $t+1$.
Traditional optical flow algorithms estimate this field using variational formulations or differential constraints.
Recently, deep learning–based models, such as RAFT~\cite{teed2020raft}, GMFlow~\cite{DBLP:conf/cvpr/XuZ0RT22}, and Sea-RAFT~\cite{DBLP:conf/eccv/WangLD24}, achieve state-of-the-art accuracy and robustness across diverse benchmarks.


\begin{figure*}
    \centering
    \includegraphics[width=0.96\textwidth]{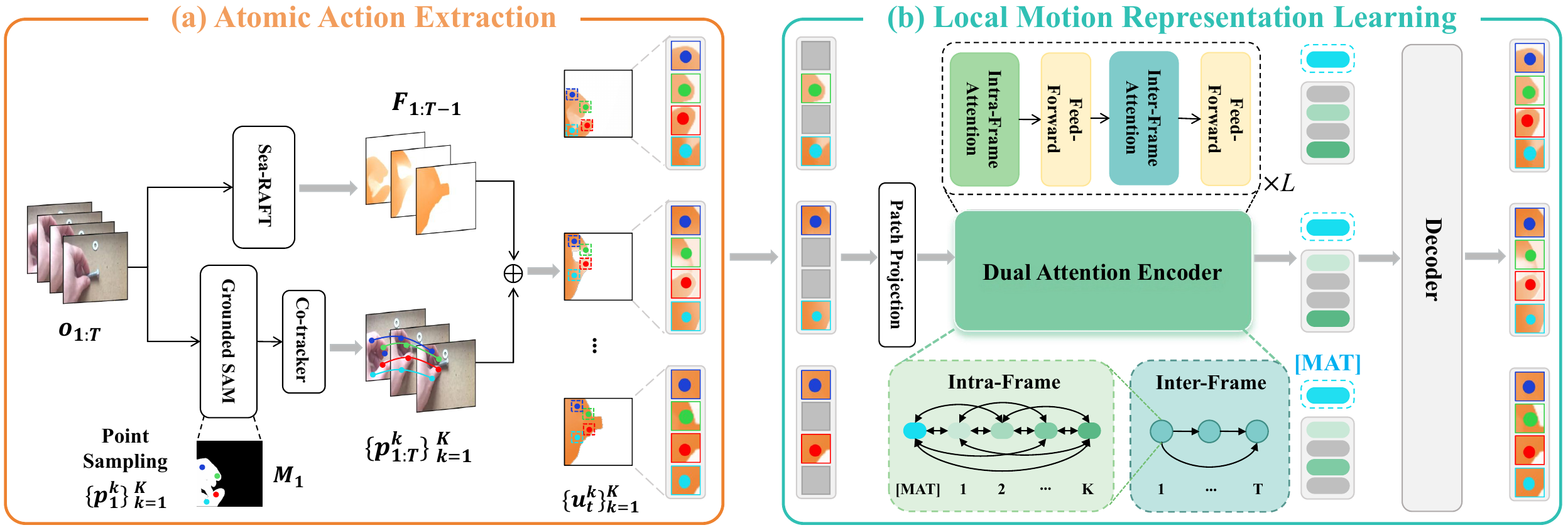}
    \caption{\textbf{Two sub-processes of the Deconstruct phase.} 
    \textbf{(a) Atomic Action Extraction:} The pipeline for extracting Atomic Actions ($u_t^k$) using Sea-RAFT, Grounded SAM, and Co-tracker.
    \textbf{(b) Local Motion Representation Learning:} The architecture of the Dual Attention Encoder (DAE), illustrating Patch Projection and Intra-/Inter-Frame Attention mechanisms.}
    \label{fig: extraction and representation}
\end{figure*}

\section{Method}
\label{sec:method}

\subsection{Overview}
\label{ssec:overview}
To learn morphology-agnostic local motion patterns for effective cross-domain transfer, we propose a novel Deconstruct-Recompose Paradigm (DRP).
The Deconstruct phase (Section~\ref{ssec:deconstruct}) includes two key sub-processes: Atomic Action extraction (Section~\ref{ssec:atomic_action_extraction}), where complex global motions are broken down into local components, termed Atomic Actions; 
and local motion representation learning (Section~\ref{ssec:learning_representations}), where we introduce a Dual Attention Encoder (DAE) to capture spatiotemporal relationships of these Atomic Actions.
The Recompose phase (Section~\ref{ssec:recompose}) composes these local motion representations to align with the specific task.
Finally, Section~\ref{ssec:drp_training} provides details of the DRP training schedule.
As shown in Figure~\ref{fig:architecture}, DRP uses a two-stage pre-training and fine-tuning framework.
\subsection{Deconstructing Local Motion Representations}
\label{ssec:deconstruct}
As shown in Figure~\ref{fig: extraction and representation}, the Deconstruct phase first breaks down complex global motion into a set of morphology-agnostic local components, termed Atomic Actions.
We then learn local motion representations to capture the spatiotemporal relationships of these Atomic Actions.

\subsubsection{Atomic Action Extraction}
\label{ssec:atomic_action_extraction}

As illustrated in Figure~\ref{fig: extraction and representation}(a), we design a pipeline to extract Atomic Actions.

First, given a video sequence $o_{1:T}$, we leverage a pre-trained Sea-RAFT model~\cite{DBLP:conf/eccv/WangLD24} to compute dense optical flow fields $F_{1:T-1}$ between consecutive frames, representing the global motion.

Second, to focus on meaningful local components and avoid static background regions, we identify and track multiple motion-salient keypoints.
In practice, we utilize Grounded SAM~\cite{DBLP:journals/corr/abs-2401-14159} to segment individuals and obtain the foreground mask $M_1$ for the first frame. 
Then, $K$ candidate points are randomly sampled within $M_1$ and tracked across the video using Co-tracker~\cite{karaev23cotracker}.
To ensure these points represent salient motion, we filter out those with low motion variance over time and resample, yielding the filtered keypoints $\{p_1^k\}_{k=1}^K$. 
Co-tracker then tracks these filtered keypoints again, yielding stable trajectories $\{p_{1:T}^k = (x_{1:T}^k, y_{1:T}^k)\}_{k=1}^{K}$ that consistently correspond to the same local physical region.



Finally, at each time step $t$, we crop a $P \times P$ local optical flow patch $u_t^k \in \mathbb{R}^{P \times P \times 2}$ from the global flow field $F_{t}$, centered at each tracked keypoint $p_t^k$. This flow patch $u_t^k$ is defined as our \textit{Atomic Action}.

\begin{figure*}
    \centering
    \includegraphics[width=\textwidth]{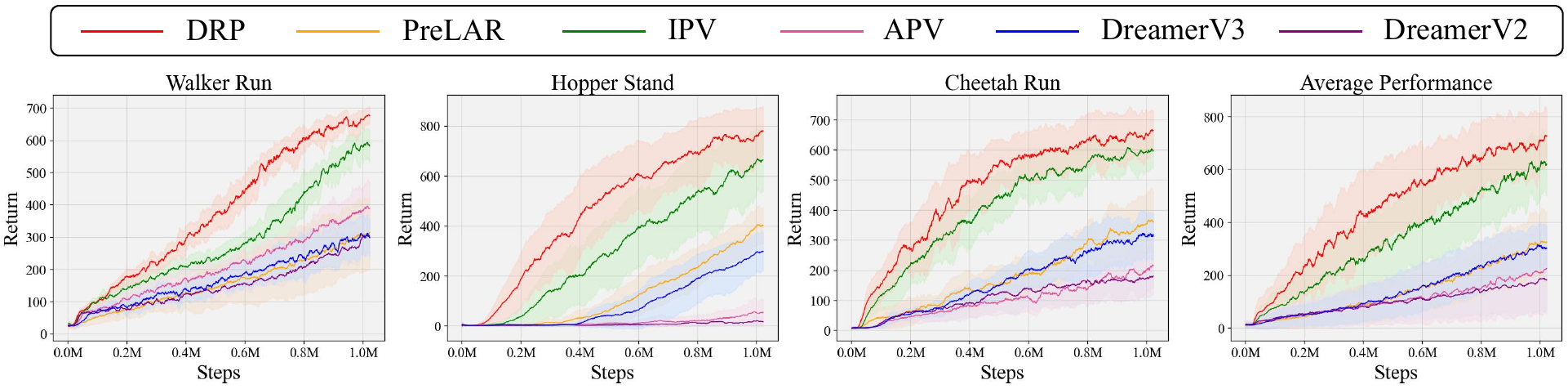}
    \caption{\textbf{Learning curves on DMControl Remastered.} Comparison of our method (DRP) against baselines on three locomotion tasks and their average. Solid lines represent the mean episode return, and shaded areas denote the standard deviation over five random seeds.}
    \label{fig:DMCR}
\end{figure*}

\subsubsection{Local Motion Representations Learning}
\label{ssec:learning_representations}

To learn a local motion representation that captures the spatiotemporal relationships among the extracted Atomic Actions, we propose a novel Dual-Attention Encoder (DAE).
The DAE incorporates two types of attention (Intra-Frame and Inter-Frame), which separately model spatial and temporal relationships, as shown in Figure~\ref{fig: extraction and representation}(b)


\textbf{Dual Attention Encoder Architecture.}
The DAE is a Transformer-based encoder~\cite{DBLP:conf/nips/VaswaniSPUJGKP17}.
Before encoding, each extracted Atomic Action $u_t^k$ is tokenized into $\tau_t^k$ by combining three types of embeddings:
(1) a content embedding obtained via Patch Projection of $u_t^k$;
(2) a coordinate embedding derived from the tracked keypoint  position $p_t^k$;
(3) a temporal embedding corresponding to timestep $t$.
A reserved learnable Motion Aggregation Token `[MAT]' is prepended to the token sequence of each frame for the subsequent Recompose phase.


The resulting token set $\{\text{[MAT]}, \tau_t^1, \dots, \tau_t^K\}_{t=1}^{T}$ is processed by $L$ stacked Dual-Attention Blocks, each comprising two complementary attention branches:
(1) \textit{Intra-Frame Attention:} performs self-attention among all tokens within the same timestep $t$, enabling the model to capture spatial relationships among different local motion parts while maintaining spatial consistency.
(2) \textit{Inter-Frame Attention:} applies causal self-attention across timesteps for each local part $k$, allowing the model to capture the temporal relationships of local motion trajectories.

\textbf{Learning Objectives.}
The DAE is trained with a Masked AutoEncoder (MAE)~\cite{DBLP:conf/cvpr/HeCXLDG22} reconstruction objective.
As shown in Figure~\ref{fig: extraction and representation}(b), we randomly mask a subset of tokens from the input sequence, and a decoder reconstructs the original patches from the visible tokens.
This objective encourages the DAE to model the spatiotemporal relationships necessary to infer the masked Atomic Actions.

\subsection{Recomposing Local Motion Representations}
\label{ssec:recompose}

The Recompose phase composes the local motion representations from the Deconstruct phase (Section~\ref{ssec:deconstruct}) by learning a latent dynamics model.

As shown in Figure~\ref{fig:architecture}, the reserved `[MAT]' token aggregates local motion representations at timestep $t$ into an aggregated action representation $a_t^{\text{agg}}$.
Based on $a_t^{\text{agg}}$, we learn a latent dynamics model. 
Given the current observation $o_t$ and the previous aggregated action representation $a_{t-1}^{\text{agg}}$, the dynamics model infers the latent state $z_t$:
\begin{equation}
z_t \sim q_\theta(z_t \mid z_{t-1}, a_{t-1}^{\text{agg}}, o_t).
\end{equation}
Following Dreamer~\cite{DBLP:conf/iclr/HafnerLB020}, the model is trained by minimizing an objective that combines image reconstruction and KL divergence regularization:
\begin{equation}
\begin{aligned}
&\mathcal{L}_{\text{dyn}}  = \mathbb{E}_{q_\theta} \Big[ \sum_{t=1}^T \Big( - \ln p_\theta (o_t | z_t) + \beta_z \mathcal{L}_z \Big) \Big],\\
&\mathcal{L}_z  = \mathrm{KL} \left[ q_\theta(z_t | z_{t-1}, a_{t-1}^{\text{agg}},o_t) \parallel p_\theta(\hat{z}_t | z_{t-1}, a_{t-1}^{\text{agg}}) \right],
\end{aligned}
\label{eq:dyn_loss}
\end{equation}
where $\beta_z$ is a hyperparameter.
The learning objective of the latent dynamics model ensures that the learned representations possess dynamic semantics.


\subsection{DRP Training Schedule}
\label{ssec:drp_training}
We train our proposed Deconstruct-Recompose Paradigm (DRP) within a two-stage pre-training and fine-tuning framework, as illustrated in Figure~\ref{fig:architecture}.

\textbf{Pre-training.}
During pre-training, we execute the Deconstruct and Recompose processes sequentially to learn transferable local motion representations, as shown in Figure~\ref{fig:architecture}(a).
First, the Deconstruct process is performed through Atomic Action extraction and local motion representation learning with DAE.
Second, we perform the Recompose process by training a latent dynamics model, which leverages the reserved `[MAT]' token to compose the learned local motion representations.

\begin{figure*}
    \centering
    \includegraphics[width=\textwidth]{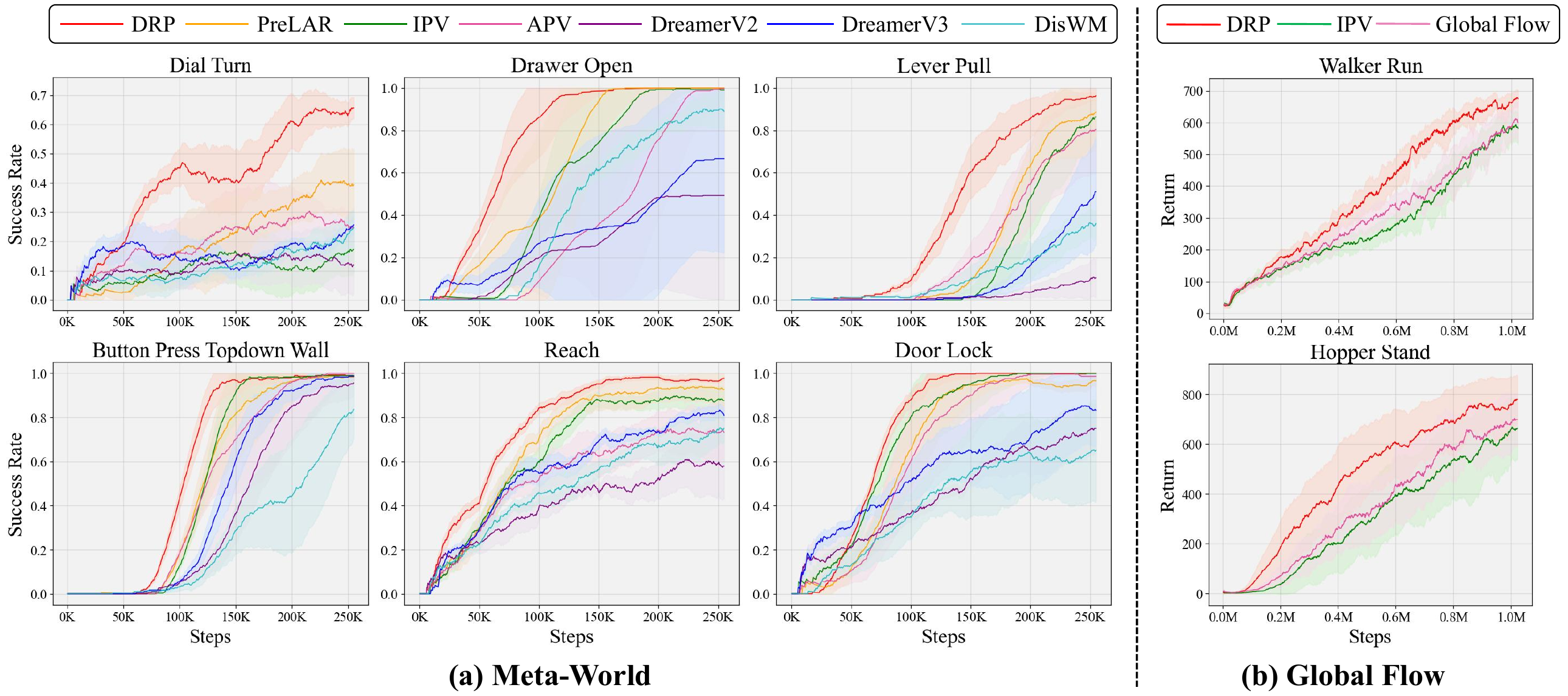}
    \caption{\textbf{Results on Meta-World (left) and ablation on the importance of motion deconstruction (right).} (a) Learning curves of our method (DRP) compared to baselines across six tasks in Meta-World, based on the average success rate over five runs. (b) Comparison between global flow modeling and our local motion representation modeling on DMCR tasks.}
    \label{fig:metaworld and global flow}
\end{figure*}

\textbf{Fine-tuning.}
As shown in Figure~\ref{fig:architecture}(b), the fine-tuning stage progressively adapts the Deconstruct–Recompose Processes.
We also map local motion representations to the agent’s specific action space to learn the downstream policy.
First, we adapt the Deconstruct process by slowly fine-tuning the pre-trained DAE.
This allows DAE to preserve learned local motion representations while progressively incorporating new patterns not included in the pre-training dataset.
Second, we adapt the Recompose process by fine-tuning the pre-trained latent dynamics model.
As a result, the `[MAT]' learns to selectively recompose the local motion representations to match the downstream agent.

Furthermore, to map these representations to the downstream action space,  we introduce an adapter and an Action-Specific Dynamics Model, as shown in Figure~\ref{fig:architecture}(b).
The adapter serves as a bridge between the pre-trained latent dynamics model and the Action-Specific Dynamics Model. 
This new Action-Specific Dynamics Model is conditioned on the agent's real action signals ($a_t$).
Specifically, it updates its agent-specific state $s_t$ by taking two inputs: the agent's previous action $a_{t-1}$, and the latent state $z_t$ from the pre-trained model after it has been mapped by the adapter:
\begin{equation} 
s_t \sim q_\phi(s_t | s_{t-1}, a_{t-1}, z_t) .
\end{equation}
The objective function for the Action-Specific Dynamics Model includes image reconstruction loss, reward prediction loss, and a KL divergence term:
\begin{equation}
\begin{aligned}
\mathcal{L}_{\text{action}} = \mathbb{E}_{q_\phi, q_\theta} \Big[ \sum_{t=1}^T \Big( - \ln p_\theta (o_t | s_t, c) - \beta_r \ln p_\varphi (r_t | s_t) \\ + \beta_s \mathrm{KL} \left[ q_\phi(s_t | s_{t-1},a_{t-1}, z_t) \parallel p_\phi(\hat{s}_t | s_{t-1}, a_{t-1}) \right] \Big) \Big],
\end{aligned}
\end{equation}
where $\beta_r$ and $\beta_s$ are hyperparameters, and $c$ is a context variable as defined in IPV~\cite{DBLP:conf/nips/0001MDL23}.
This architecture tailors the pre-trained knowledge to the specific downstream agent, accelerating downstream policy learning.

\section{Experiments}
\label{sec:experiments}

We conduct extensive experiments to validate the effectiveness of our proposed Deconstruct-Recompose Paradigm (DRP). 
Our evaluation is performed on two challenging downstream benchmarks: the robotic locomotion suite, DMControl Remastered (DMCR)~\cite{DBLP:journals/corr/abs-2010-06740}, and the robotic manipulation benchmark, Meta-World~\cite{DBLP:conf/corl/YuQHJHFL19}. 
Our experiments aim to answer the following key questions:



\begin{itemize}
    \item \textbf{Q1:} Is our Deconstruct-Recompose paradigm broadly effective for various downstream tasks?
    \item \textbf{Q2:} Is deconstructing global motion into local Atomic Actions crucial for effective knowledge transfer?
    \item \textbf{Q3:} Is the dual-attention mechanism necessary for learning robust local motion representations?
    \item \textbf{Q4:} Are the improvements on downstream tasks due to the pre-trained local motion representations?
    \item \textbf{Q5:} What benefits does the pre-trained local motion representation bring for cross-domain transfer?
\end{itemize}




\textbf{Pre-training datasets.} 
Consistent with IPV~\cite{DBLP:conf/nips/0001MDL23} and PreLAR~\cite{DBLP:conf/eccv/ZhangKSC24}, we use the Something-Something-V2 (SSV2) dataset~\cite{DBLP:conf/iccv/GoyalKMMWKHFYMH17} for pre-training.

\textbf{Baselines.}
We compare our method (DRP) with two state-of-the-art Model-Based Reinforcement Learning (MBRL) algorithms and four unsupervised pre-training approaches in RL:
1) \textbf{DreamerV2}~\cite{DBLP:conf/iclr/HafnerL0B21} is the first MBRL algorithm to reach human-level performance on the Atari benchmark.
2) \textbf{DreamerV3}~\cite{hafner2025dreamerv3} is currently a powerful MBRL algorithm, outperforming specialized approaches across more than 150 diverse tasks.
3) \textbf{APV}~\cite{DBLP:conf/icml/SeoLJA22} pre-trains an action-free world model using unlabeled video data from a different domain: RLBench~\cite{DBLP:journals/ral/JamesMAD20}. 
4) \textbf{IPV}~\cite{DBLP:conf/nips/0001MDL23} introduces Contextualized World Models in pre-training with rich in-the-wild videos.
5) \textbf{PreLAR}~\cite{DBLP:conf/eccv/ZhangKSC24} introduces a learnable action representation to leverage action-free videos for pre-training world models.
6) \textbf{DisWM}~\cite{wang2025disentangled} transfers the disentanglement capability of the pre-trained model to the downstream world model through latent distillation.
For a fair comparison, all unsupervised pre-training baselines are pre-trained on the SSV2 dataset.

\begin{figure*}
    \centering
    \includegraphics[width=\textwidth]{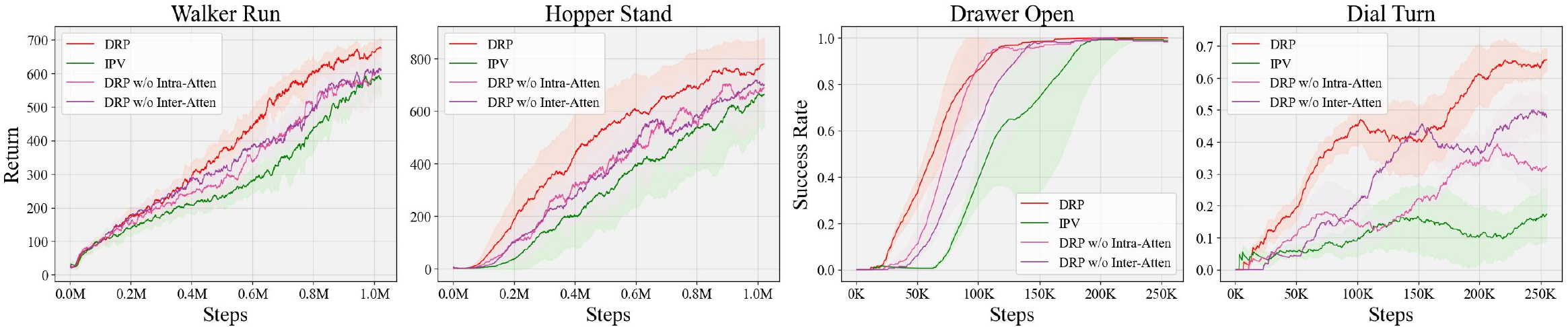}
    \caption{\textbf{Ablation study on DA-MAE architectural components.} We analyze the necessity of our dual-attention mechanisms. The results compare our model against variants removing Intra-Frame Attention (w/o Intra-Atten) or Inter-Frame Attention (w/o Inter-Atten).}
    \label{fig:ablation_study}
\end{figure*}

\subsection{Evaluation on DMControl Remastered}
\label{ssec:eval_dmcr}

To evaluate the effectiveness of our proposed Deconstruct–Recompose Paradigm (DRP) on downstream robotic locomotion tasks, we conduct experiments on the challenging simulated robotics benchmark DMControl Remastered (DMCR)~\cite{DBLP:journals/corr/abs-2010-06740}.
DMCR is designed to measure generalization in continuous control by featuring randomly generated, complex graphics. 


Following IPV~\cite{DBLP:conf/nips/0001MDL23}, we evaluate on three distinct locomotion tasks: ``Walker Run'', ``Hopper Stand'', ``Cheetah Run''. 
Figure~\ref{fig:DMCR} presents the learning curves for individual tasks and their aggregated average, comparing our method against strong baselines. 
As shown in Figure~\ref{fig:DMCR}, our method consistently achieves significant improvements in both sample efficiency and asymptotic performance across all tasks.

\subsection{Evaluation on Meta-World}
\label{ssec:eval_metaworld}


To further investigate the broad effectiveness of our method (DRP), we extend our evaluation to robotic manipulation tasks using the widely adopted Meta-World benchmark~\cite{DBLP:conf/corl/YuQHJHFL19}.
Following the IPV~\cite{DBLP:conf/nips/0001MDL23} and PreLAR~\cite{DBLP:conf/eccv/ZhangKSC24}, we evaluate our method on the same six tasks.

The learning curves are presented in Figure~\ref{fig:metaworld and global flow}(a). 
The results demonstrate that our method (DRP) achieves state-of-the-art performance across all six tasks.
We observe significant improvements in asymptotic performance on the ``Dial Turn'' and ``Lever Pull'' tasks. 
Notably, our method successfully masters the ``Dial Turn'' task, which is considered particularly challenging, whereas baselines struggle.
Furthermore, our approach significantly improves sample efficiency on the ``Drawer Open'', ``Button Press Topdown wall'', ``Reach'', and ``Door Lock'' tasks. 

Combined with the DMCR results, our DRP consistently achieves superior performance compared with both video prediction methods (IPV~\cite{DBLP:conf/nips/0001MDL23}, APV~\cite{DBLP:conf/icml/SeoLJA22}, and DisWM~\cite{wang2025disentangled}) and latent action modeling methods (PreLAR~\cite{DBLP:conf/eccv/ZhangKSC24}).
These experiments provide clear evidence supporting \textbf{Q1}, demonstrating that our DRP is broadly effective across diverse downstream tasks.

\subsection{Importance of Motion Deconstruction}
\label{ssec:ablation_decomposition}

We now address question \textbf{Q2}: \textit{Is deconstructing global motion into local Atomic Actions crucial for cross-domain transfer?} 
To validate this hypothesis, we conduct a key ablation study comparing our local motion modeling method with the conventional global motion modeling paradigm.

To this end, we design a baseline variant termed ``Global Flow'', representing the global modeling approach.
It replaces our Deconstruct pipeline with a convolutional encoder that processes the \textit{entire} global optical flow into a single global embedding to condition the dynamics model. 
For a fair comparison, both models are pre-trained on the SSV2 dataset and fine-tuned on DMCR tasks (``Walker Run'' and ``Hopper Stand''), differing \textit{only} in their motion representation strategy (local \textit{vs.} global).

As shown in Figure~\ref{fig:metaworld and global flow}(b), our DRP method significantly outperforms the ``Global Flow'' variant in both sample efficiency and asymptotic return. 
Notably, ``Global Flow'' fails to show an advantage over IPV~\cite{DBLP:conf/nips/0001MDL23}, which also relies on global image modeling.

This result provides strong evidence that global modeling is tightly coupled with morphology, which hinders transfer across domains.
In contrast, the local motion representations learned through deconstructing global motion enable effective cross-domain transfer.

\begin{figure}
    \centering
    \includegraphics[width=0.50\textwidth]{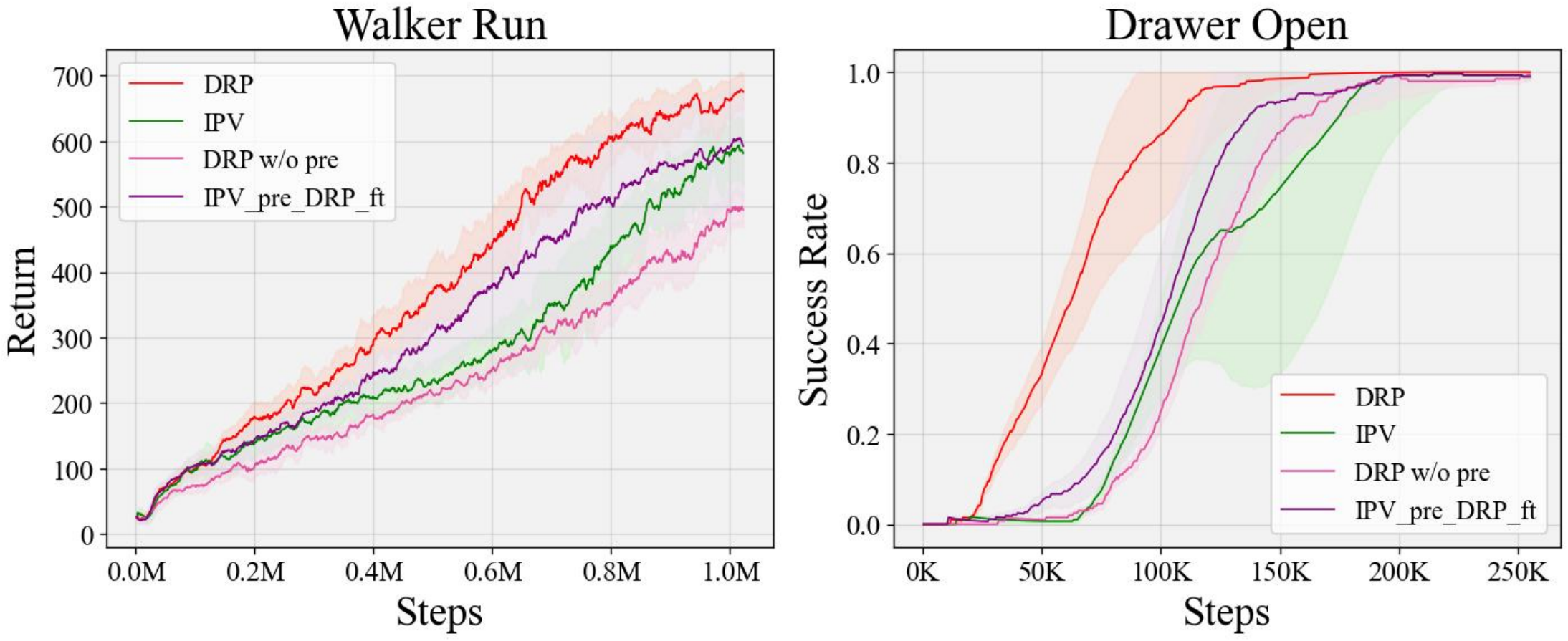}
    \caption{\textbf{Ablation on pre-training effectiveness.} Removing our pre-training or replacing it with IPV’s leads to clear performance degradation, confirming the superiority of our pre-training.}
    \label{fig:ablation_pretrain}
\end{figure}

\subsection{Ablation Study}
\label{ssec:ablation_architecture}

\subsubsection{Dual-Attention Mechanism in DAE}
We first address \textbf{Q3}: \textit{Is the dual-attention mechanism necessary for learning local motion patterns?} We conduct an ablation study by removing the \textit{Intra-Frame Attention} and \textit{Inter-Frame Attention} components from the DAE, respectively. 
We evaluate these variants on two tasks in DMCR (``Walker Run'', ``Hopper Stand'') and two tasks in Meta-World (``Drawer Open'', ``Dial Turn'').

Figure~\ref{fig:ablation_study} presents the results. 
Both ablated variants show significant performance degradation compared to DRP. 
This confirms that both mechanisms are essential: Intra-Frame Attention is crucial for modeling spatial relationship among different parts; Inter-Frame Attention is necessary for capturing their individual temporal relationship.


\subsubsection{Transfer Gain Attribution}
Next, we address \textbf{Q4}: \textit{Are the improvements on downstream tasks due to the pre-trained local motion representations?} We conduct two critical comparisons:
(1) w/o pre-training: We train our model from scratch (``DRP w/o pre'') to validate the benefit of pre-training.
(2) vs. Global Pre-training: To isolate the advantage of our pre-training based on local motion representations, we compare it with a variant (``IPV\_pre\_DRP\_ft'').
This variant utilizes a pre-trained model from a global modeling method, IPV~\cite{DBLP:conf/nips/0001MDL23}, but employs our identical fine-tuning framework.

As shown in Figure~\ref{fig:ablation_pretrain}, our full method (DRP) significantly outperforms ``DRP w/o pre'', demonstrating that pre-training provides valuable prior knowledge and accelerates downstream policy learning.
Importantly, our method also shows a clear performance improvement compared to ``IPV\_pre\_DRP\_ft''.
In addition, we ablate the pre-trained dynamics model.
As shown in Table~\ref{table:dynamic_ablation_variation}, removing our dynamics model (``DRP w/o dyn.'') causes a massive performance degradation. 
In contrast, removing IPV's dynamics (``IPV w/o dyn.'') model has a minimal impact. 
These results exhibit that our pre-trained local motion representations have learned transferable motion patterns, while the baselines primarily focus on static representations.

\begin{table}
\setlength\tabcolsep{2pt}
\caption{\textbf{Ablation study on pre-trained dynamics model.} We report the mean and standard deviation on DMCR tasks.}
\centering
\resizebox{0.32\textwidth}{!}{%
\begin{tabular}{l | c c}
    \toprule
    Method & Walker Run & Hopper Stand \\
    \midrule
     DRP & 681 $\pm$ 39 & 796$\pm$ 114 \\
     DRP w/o dyn. & 613 $\pm$ 45  & 708 $\pm$ 106 \\
        $\Delta_\text{dyn.}$ & \textbf{$\downarrow$10.0\%}   &
    \textbf{$\downarrow$11.1\%} \\
    \midrule
     IPV & 595 $\pm$ 67 & 634 $\pm$ 128 \\
     IPV w/o dyn. & 586 $\pm$ 61 & 628 $\pm$ 134 \\
     $\Delta_\text{dyn.}$ & \textbf{$\downarrow$1.5\%}  & \textbf{$\downarrow$0.9\%} \\
    \bottomrule
\end{tabular}%
}
\label{table:dynamic_ablation_variation}
\end{table}

\subsection{Visualization of Video Predictions}
\label{ssec:dynamics_analysis}

Finally, we address \textbf{Q5}: \textit{What benefits does the pre-trained local motion representation bring for cross-domain transfer?} 
To investigate this question, we perform open-loop video prediction in both the source domain and cross-domain settings to illustrate the results.

\begin{figure}
    \centering
    \includegraphics[width=0.47\textwidth]{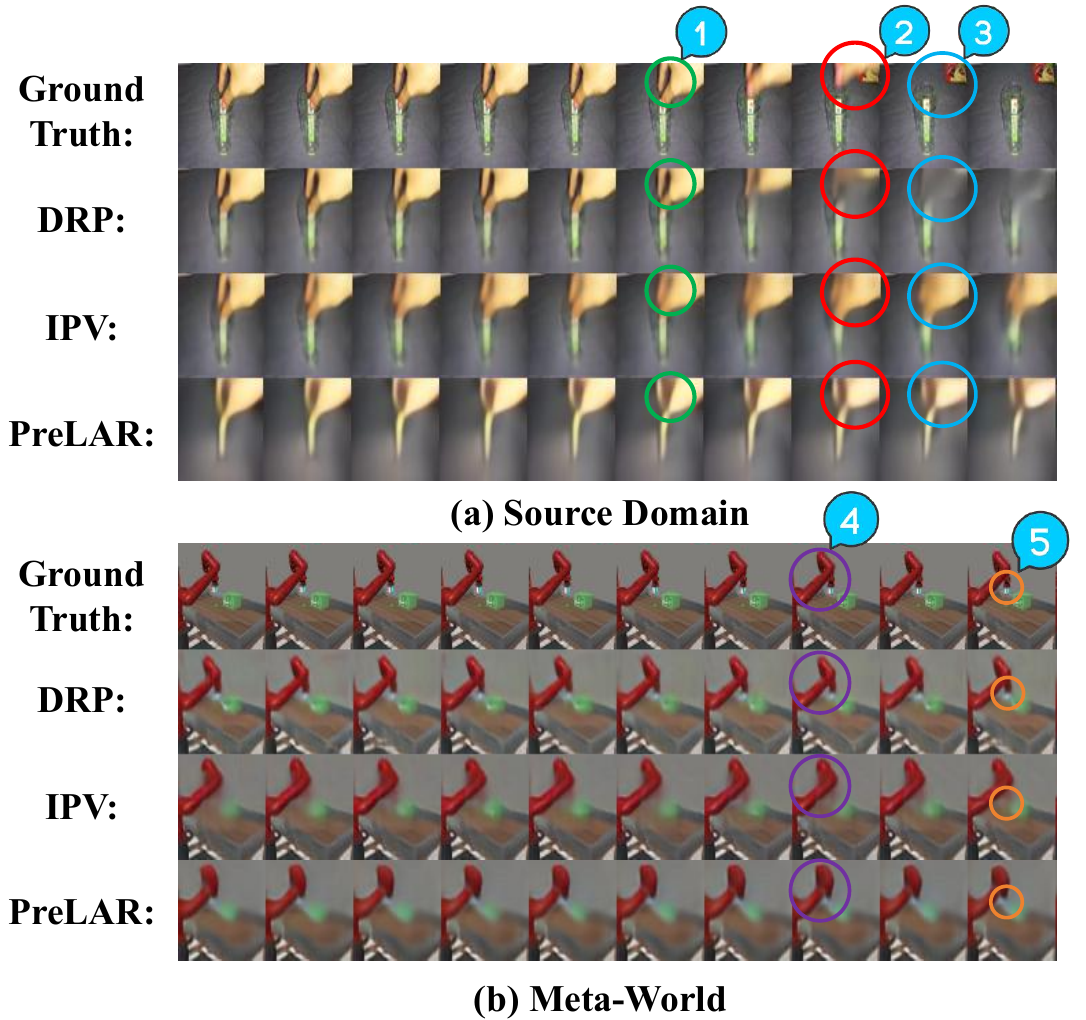}
    \caption{\textbf{Visualization of Video Predictions.} Open-loop predictions on (a) the source domain (SSV2) and (b) a zero-shot cross-domain setting (Meta-World).}
    \label{fig:video_prediction}
\end{figure}


First, we examine whether the learned local motion representations accurately capture motion patterns in the source domain.
Given the initial frames of a video, we perform open-loop video prediction using our dynamics model and compare the results with global-modeling approaches, IPV~\cite{DBLP:conf/nips/0001MDL23} and PreLAR~\cite{DBLP:conf/eccv/ZhangKSC24}.
As shown in Figure~\ref{fig:video_prediction}(a), our method predicts future frames that more accurately reflect long-horizon motion.
For example: 1) DRP more accurately captures the shape of the hand and correctly predicts the subsequent 2) movement and 3) disappearance of the hand from the scene.
In contrast, IPV and PreLAR fail to capture these details.



Second, we evaluate cross-domain transfer by performing zero-shot open-loop predictions on a robot manipulation task (Meta-World).
As shown in Figure~\ref{fig:video_prediction}(b), our DRP generates accurate future-frame predictions for previously unseen robotic agents. 
Specifically, 4) DRP more accurately preserves the shape of the robotic arm, whereas IPV and PreLAR exhibit varying degrees of arm deformation. 
In addition, 5) DRP is able to capture the robotic gripper, which is crucial for manipulation tasks, while the gripper disappears in the predictions of IPV and PreLAR.
This demonstrates that our pre-trained local motion representations facilitate capturing dynamic semantic information.

\section{Conclusion}
\label{sec:conclusion}

Video pre-training offers a promising path to improving downstream RL. The major obstacle lies in the morphological mismatch between pre-training data and agents. Nonetheless, local motion patterns exhibit cross-agent similarity.
Inspired by this insight, we propose a novel Deconstruct–Recompose Paradigm (DRP) to model such transferable local motion patterns.
In the Deconstruct phase, global motion is broken down into a set of local Atomic Actions, and a Dual-Attention Encoder (DAE) learns local motion representations from these Atomic Actions, capturing their spatiotemporal relationships.
In the Recompose phase, these local motion representations are composed using a learnable Motion Aggregation Token `[MAT]' through latent dynamics model learning.
Extensive experiments demonstrate that DRP significantly improves sample efficiency and performance across a wide range of downstream robotic control and manipulation tasks.
\section*{Acknowledgments}
This work was supported by the National Natural Science Foundation of China (No. 62576029) and the National Natural Science Foundation of China ( No. 62536001).
{
    \small

}
\clearpage
\setcounter{page}{1}
\maketitlesupplementary

\section{Details of Pre-training and Fine-tuning}
\label{appendix:optimization}
This section provides the detailed optimization objectives used in both the Pre-training and Fine-tuning stages.
These objectives correspond directly to the DRP training schedule described in Section~\ref{ssec:drp_training} of our main manuscript.
The complete pre-training and fine-tuning algorithms are provided in Algorithm~\ref{alg:pretrain} and Algorithm~\ref{alg:finetune}, respectively.

\subsection{Pre-training Objectives}

During pre-training, DRP optimizes two objectives: a Masked AutoEncoder (MAE) reconstruction objective for deconstructing global motion into local motion representations, and a latent dynamics model objective for recomposing local motion representations. 
The full pre-training stage is summarized in Algorithm~\ref{alg:pretrain}.



\paragraph{MAE Reconstruction Objective.}
To learn local motion representations that capture the spatiotemporal relationships among Atomic Actions, we train the Dual-Attention Encoder (DAE) using a Masked Autoencoder (MAE) reconstruction objective.
During pre-training, we randomly mask a subset of Atomic Action tokens and require the decoder to reconstruct only the corresponding masked Atomic Actions from the visible tokens:
\begin{equation}
\mathcal{L}_{\text{MAE}}
=
\frac{1}{|\mathcal{M}|}
\sum_{i \in \mathcal{M}}
\left\|
    \hat{u}_i - u_i
\right\|_2^2 ,
\end{equation}
where $\mathcal{M}$ denotes the set of masked Atomic Actions, $u_i$ represents the original Atomic Action patch, and $\hat{u}_i$ is the reconstructed patch. 
This objective encourages the DAE to model the spatiotemporal relationships necessary to infer the masked Atomic Actions.

\paragraph{Latent Dynamics Objective.}
The latent dynamics model learns to recompose local motion representations through the aggregated action representation $a_t^{\text{agg}}$. 
Following Dreamer, the latent dynamic model is trained with an image reconstruction term and a KL regularization term:
\begin{equation}
\label{eq:dyn_loss_appendix}
\begin{aligned}
\mathcal{L}_{\text{dyn}}
&= \mathbb{E}_{q_\theta} \Bigg[
    \sum_{t=1}^T
    \Big(
        -\ln p_\theta(o_t \mid z_t)
        + \beta_z \, \mathcal{L}_z
    \Big)
\Bigg], \\
\mathcal{L}_z 
&= \mathrm{KL}\!\left[
        q_\theta(z_t \mid z_{t-1}, a_{t-1}^{\text{agg}}, o_t)
        \,\Big\|\,
        p_\theta(\hat{z}_t \mid z_{t-1}, a_{t-1}^{\text{agg}})
    \right].
\end{aligned}
\end{equation}
This objective ensures that the recomposed motion representation acquires dynamic semantics.

\begin{algorithm}[t]
\caption{DRP Pre-training}
\label{alg:pretrain}
\begin{algorithmic}[1]

\State Initialize parameters $\zeta$ of DAE, $\theta$ of latent dynamics model, image encoder, and decoder randomly
\State Load unlabeled video dataset $\mathcal{D}$

\For{every iteration}
    \State Randomly sample videos $\{o_{1:T}\} \sim \mathcal{D}$

    \textcolor[rgb]{0.7,0.7,0.7}{\textbf{\# Atomic Action Extraction}}
    \State Compute optical flow $\{F_{1:T-1}\}$ using Sea-RAFT
    \State Segment foreground mask $M_1$ via Grounded SAM
    \State Track $K$ keypoints $\{p_{1:T}^k\}$ using Co-tracker
    \State Extract Atomic Action patches $\{u_t^k\}$

    \textcolor[rgb]{0.7,0.7,0.7}{\textbf{\# Local Motion Representation Learning}}
    \State Tokenize $u_t^k$ into $\tau_t^k$ and prepend `[MAT]'
    \State Encode tokens via DAE to obtain local motion representations

    \textcolor[rgb]{0.7,0.7,0.7}{\textbf{\# Recompose via Latent Dynamics Model}}
    \State Obtain aggregated representation $a_t^{agg}$ from `[MAT]'
    \State Infer the latent state $z_t$:
        \[
        z_t \sim q_\theta(z_t \mid z_{t-1}, a_{t-1}^{agg}, o_t).
        \]
    \State Update latent dynamic model by minimizing $\mathcal{L}_{dyn}$ in Eq.~\eqref{eq:dyn_loss_appendix}.
\EndFor

\end{algorithmic}
\end{algorithm}

\subsection{Fine-tuning Objectives}


The Fine-tuning stage jointly fine-tunes the DAE and the latent dynamics model, and maps the recomposed motion representations into the downstream agent's action space to accelerate policy learning. 
The overall objective consists of three components: 
(1) MAE fine-tuning of the DAE, 
(2) fine-tuning of the latent dynamics model, and 
(3) the Action-Specific Dynamics Model objective, as detailed in Algorithm~\ref{alg:finetune}.

\paragraph{MAE Fine-tuning for DAE.}
The DAE is slowly fine-tuned using the same MAE objective as in pre-training:
\begin{equation}
\label{ft_mae}
\mathcal{L}_{\text{MAE}}^{\text{ft}} = \mathcal{L}_{\text{MAE}}.
\end{equation}
This retains the pre-trained transferable local motion representations while allowing the model to gradually incorporate new ones specific to the downstream agent (e.g., rigid-joint motions).

\paragraph{Latent Dynamics Fine-tuning.}
The pre-trained latent dynamics model is slowly fine-tuned so that the Motion Aggregation Token `[MAT]' can selectively recompose the local motion representations to match the downstream agent:
\begin{equation}
\label{ft_dynamic}
\mathcal{L}_{\text{dyn}}^{\text{ft}} = \mathcal{L}_{\text{dyn}},
\end{equation}
with the same form as in Eq.~\eqref{eq:dyn_loss_appendix}.

\paragraph{Action-Specific Dynamics Model Objective.}
We further introduce an Adapter together with an Action-Specific Dynamics Model to map the recomposed local motion representations into the downstream agent’s action space. 
The Adapter bridges the pre-trained latent dynamics model and the Action-Specific Dynamics Model.

The optimization objective of the Action-Specific Dynamics Model is:
\begin{equation}
\begin{aligned}
\label{action_dynamic}
\mathcal{L}_{\text{action}} = \mathbb{E}_{q_\phi, q_\theta} \Big[ \sum_{t=1}^T \Big( - \ln p_\theta (o_t | s_t, c) - \beta_r \ln p_\varphi (r_t | s_t) \\ + \beta_s \mathrm{KL} \left[ q_\phi(s_t | s_{t-1},a_{t-1}, z_t) \parallel p_\phi(\hat{s}_t | s_{t-1}, a_{t-1}) \right] \Big) \Big],
\end{aligned}
\end{equation}
where $s_t$ is the agent-specific state, $z_t$ is the latent state of the pre-trained latent dynamics model.
$c$ is a context variable adopted from IPV~\cite{DBLP:conf/nips/0001MDL23}, computed by encoding randomly sampled frames from the video.
This context variable $c$ captures static context information (e.g., backgrounds and textured appearance).

\begin{algorithm}[t]
\caption{DRP Fine-tuning}
\label{alg:finetune}
\begin{algorithmic}[1]
\State Load pre-trained parameters $\zeta$ of DAE, $\theta$ of latent dynamics model, image encoder and decoder
\State  Initialize parameters $\phi$ of Action-Specific Dynamics, $\omega$ of Adapter, and $\varphi$ of reward predictors randomly
\State  Initialize parameters $\psi$ of actor $\pi_{\psi}(a|s)$ and $\xi$ of critic $v_{\xi}(s)$
\State Initialize replay buffer $\mathcal{B}$

\For{every iteration}
    \State \textcolor[rgb]{0.7,0.7,0.7}{\textbf{\# Collect Transitions}}
    \State Get state $z_t \sim q_\theta(z_t \mid z_{t-1}, a_{t-1}^{\text{agg}}, o_t)$, $s_t \sim q_{\phi}\bigl(s_t \mid s_{t-1}, a_{t-1}, z_t\bigr)$
    \State Get action $a_t \sim \pi_\psi(a_t \mid s_t)$
    \State Add transition $\{o_t, a_t, r_t\}$ to replay buffer $\mathcal{B}$
    \State \textcolor[rgb]{0.7,0.7,0.7}{\textbf{\# Deconstruct Fine-tuning}}
    \State Extract Atomic Actions $\{u_t^k\}$
    \State Fine-tune the DAE by minimizing $\mathcal{L}^{\text{ft}}_{\text{MAE}}$ in Eq.~\eqref{ft_mae}

    \State \textcolor[rgb]{0.7,0.7,0.7}{\textbf{\# Recompose Fine-tuning}}
    \State Compute aggregated representation $a_t^{agg}$
    \State Fine-tune latent dynamics model by minimizing $\mathcal{L}^{\text{ft}}_{\text{dyn}}$ in Eq.~\eqref{ft_dynamic}

    \State \textcolor[rgb]{0.7,0.7,0.7}{\textbf{\# Action-Specific Dynamics Model}}
    \State Compute latent state $z_t$ of latent dynamics model
    \State Learn Action-Specific Dynamics Model by minimizing $\mathcal{L}_{\text{action}}$ in Eq.~\eqref{action_dynamic}

    \State \textcolor[rgb]{0.7,0.7,0.7}{\textbf{\# Policy Learning}}
    \State Imagine future latent rollouts $\{\hat{s}_\tau, \hat{a}_\tau, \hat{r}_\tau\}_{\tau=t}^{t+H}$ using the Action-Specific Dynamics Model and actor
    \State Compute $\lambda$-return $V_\tau^\lambda$ in Eq.~\eqref{lamda_return}
    \State Update critic by minimizing $\mathcal{L}_{critic}$ in Eq.~\eqref{critic_loss}
    \State Update actor by minimizing $\mathcal{L}_{actor}$ in Eq.~\eqref{actor_loss}

\EndFor

\end{algorithmic}
\end{algorithm}

\section{Policy Learning}
\label{appendix:policy_learning}

During the fine-tuning stage of DRP, policy learning is performed on top of the Action-Specific Dynamics Model.
Following Dreamer, both the actor and critic are optimized entirely using imagined trajectories generated within the latent space of the Action-Specific Dynamics Model.
This design enables highly sample-efficient reinforcement learning and allows the agent to fully leverage the transferable motion knowledge acquired during pre-training.

\subsection{Latent Imagination Rollouts}

Given the current agent-specific latent state $s_t$ and policy $\pi_\psi$, the dynamics model recursively produces an imagined trajectory:
\[
\{\hat{s}_\tau, \hat{a}_\tau, \hat{r}_\tau\}_{\tau=t}^{t+H},
\]
where $H$ denotes the imagination horizon, and \(\tau\) represents the various time steps in the imagined trajectory. 
Each imagined transition is generated by:
\begin{itemize}
    \item sampling an action $\hat{a}_\tau \sim \pi_\psi(\cdot \mid \hat{s}_\tau)$,
    \item predicting the next latent state via the transition model $p_\phi(\hat{s}_{\tau+1} \mid \hat{s}_\tau, \hat{a}_\tau)$ of Action-Specific Dynamics Model,
    \item predicting the reward $\hat{r}_\tau$ via $p_\varphi (\hat{r}_\tau \mid \hat{s}_\tau)$.
\end{itemize}

These imagined trajectories serve as the sole training data for the actor and critic, 
reducing reliance on real environment interaction.

\subsection{Critic Learning}

The critic $v_\xi$ estimates the value of latent states. 
We use the $\lambda$-return to construct multi-step bootstrapped targets:
\begin{equation}
\label{lamda_return}
V_\tau^\lambda \doteq 
\hat{r}_\tau +
\gamma
\begin{cases}
(1-\lambda)\, v_\xi(\hat{s}_{\tau+1})
+
\lambda\, V_{\tau+1}^\lambda,
& \tau < t + H, \\[4pt]
v_\xi(\hat{s}_{\tau+1}),
& \tau = t + H.
\end{cases}
\end{equation}

The critic is trained to regress the \(\lambda\)-return using a squared loss:
\begin{equation}
\label{critic_loss}
\mathcal{L}_{critic}(\xi)
=
\mathbb{E}
\left[
\sum_{\tau=t}^{t+H}
\frac{1}{2}
\left(
v_\xi(\hat{s}_\tau)
-
\mathrm{sg}(V_\tau^\lambda)
\right)^2
\right],
\end{equation}
where $\mathrm{sg}(\cdot)$ is the stop-gradient operator.

\subsection{Actor Learning}

The actor is optimized to select actions that maximize the critic’s $\lambda$-return on imagined trajectories. 
An entropy regularization term encourages sufficient exploration:
\begin{equation}
\label{actor_loss}
\mathcal{L}_{actor}(\psi)
=
\mathbb{E}
\left[
\sum_{\tau=t}^{t+H}
\left(
- V_\tau^\lambda
- \eta\, \mathcal{H} \!\left[\pi_\psi(\hat{a}_\tau \mid \hat{s}_\tau)\right]
\right)
\right],
\end{equation}
where $\eta$ controls the entropy weight.
A detailed algorithmic description of the policy learning process is provided in the \textit{Policy Learning} block of Algorithm~\ref{alg:finetune}.

\begin{figure}
    \centering
    \includegraphics[width=0.50\textwidth]{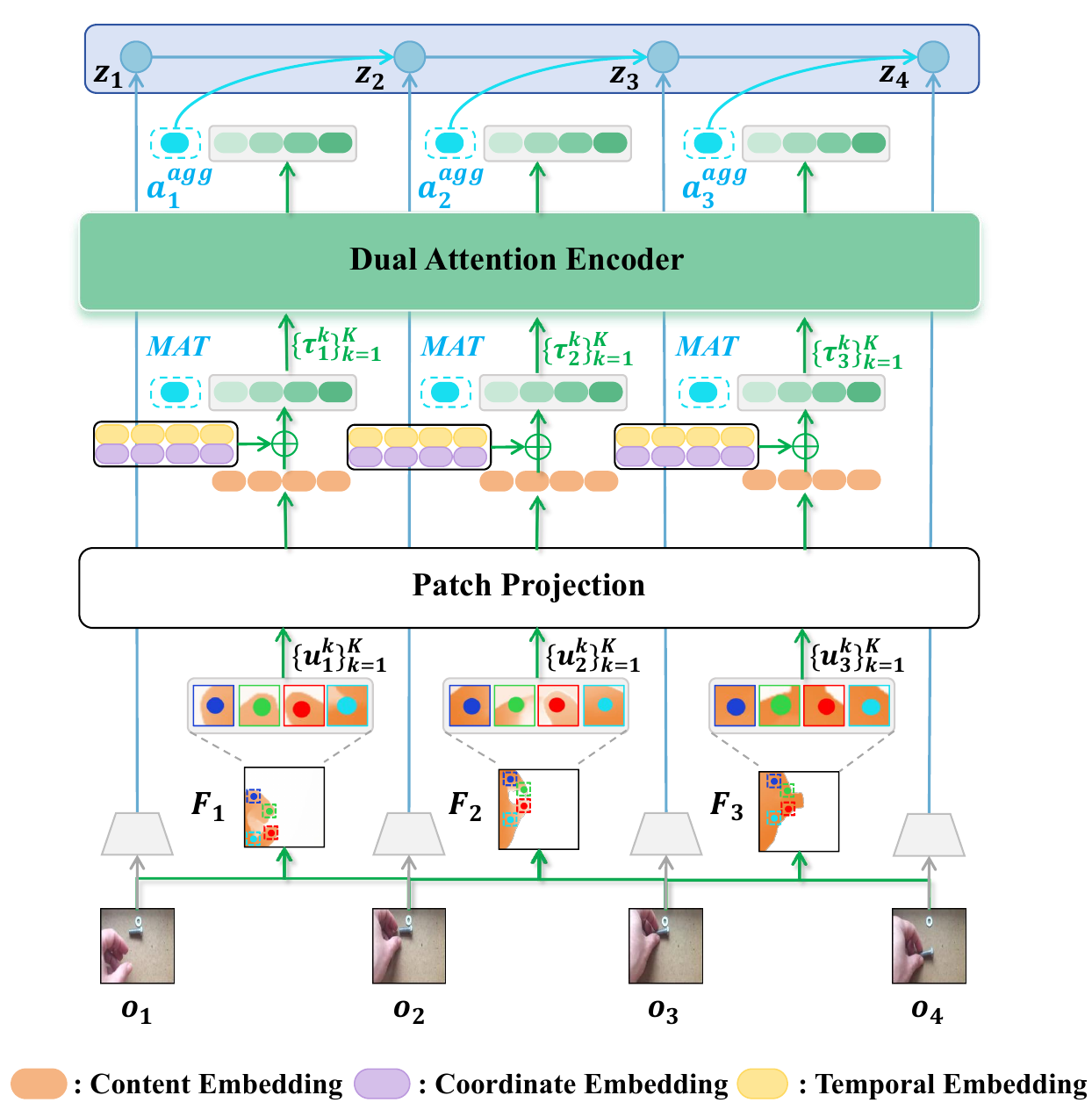}
    \caption{\textbf{Illustration of DAE Processing.} For each extracted Atomic Action, we generate a token by combining its Content Embedding with Coordinate and Temporal Embeddings. These tokens, prepended with a learnable `[MAT]' token, are processed by the DAE to produce the aggregated motion representation $a^{agg}$.}
    \label{fig:DAE_token}
\end{figure}

\section{Dual-Attention Formulation of the DAE}
\label{appendix:dae_attention}

This section provides the mathematical formulation of the Dual-Attention Encoder (DAE).  
The DAE consists of two complementary attention branches: Intra-Frame Attention and Inter-Frame Attention, 
which models the spatial and temporal relationships among Atomic Actions, respectively.


\paragraph{DAE Token Construction.}
As shown in Figure~\ref{fig:DAE_token}, for each timestep $t$, the input to the DAE is constructed from the extracted Atomic Actions. 
Given a tracked keypoint $p_t^k$, a local optical flow patch $u_t^k$ is first cropped and passed through a Patch Projection module to obtain a content embedding.
This embedding is then combined with a Coordinate Embedding and a Temporal Embedding to yield the final token embedding.
To enable aggregation during the Recompose phase, a learnable Motion Aggregation Token `[MAT]' is prepended to each frame’s token sequence. 
Thus, the complete token set at timestep $t$ is:
$\mathcal{T}_t = \{ \text{[MAT]},\; \tau_t^1,\; \tau_t^2,\;\dots,\;\tau_t^K \}$.
Stacking all tokens yields the matrix:
$X_t \in \mathbb{R}^{(K+1)\times d}$,
which serves as the input to the Dual-Attention Blocks. 
The updated `[MAT]' token after Dual-Attention produces the aggregated motion representation $a_t^{\text{agg}}$ used in the latent dynamics model.

\paragraph{Intra-Frame Attention.}
This branch models spatial relationships within each frame.
Given $X_t$, the query/key/value projections are:
\begin{equation}
Q_t = X_t W_Q^{\mathrm{intra}}, \quad 
K_t = X_t W_K^{\mathrm{intra}}, \quad
V_t = X_t W_V^{\mathrm{intra}},
\end{equation}
with $W_Q^{\mathrm{intra}}, W_K^{\mathrm{intra}}, W_V^{\mathrm{intra}} \in \mathbb{R}^{d\times d}$.
The Intra-Frame Attention output is:
\begin{equation}
\mathrm{IntraAttn}(X_t)
=
\mathrm{Softmax}\!\left(
\frac{Q_t K_t^{\top}}{\sqrt{d}}
\right)V_t.
\end{equation}
This allows the model to capture how local motion components interact spatially within the same frame.

\paragraph{Inter-Frame Attention.}
To capture temporal relationships, Inter-Frame Attention applies causal self-attention across timesteps for each local part $k$.
Given the sequence of features for the $k$-th token across time:
\[
\mathcal{S}^k = \{\tau_1^k, \tau_2^k, \dots, \tau_T^k\} \in \mathbb{R}^{T \times d},
\]
we compute:
\begin{equation}
Q^{k} = \mathcal{S}^k W_Q^{\mathrm{inter}},\quad
K^{k} = \mathcal{S}^k W_K^{\mathrm{inter}},\quad
V^{k} = \mathcal{S}^k W_V^{\mathrm{inter}},
\end{equation}
where $W_Q^{\mathrm{inter}}, W_K^{\mathrm{inter}}, W_V^{\mathrm{inter}} \in \mathbb{R}^{d\times d}$.
Causal masking is applied using $M_{\mathrm{causal}} \in \mathbb{R}^{T\times T}$ to prevent attending to future timesteps.
The Inter-Frame Attention output is:
\begin{equation}
\mathrm{InterAttn}(\mathcal{S}^k)
=
\mathrm{Softmax}\!\left(
\frac{Q^{k}(K^{k})^{\top}}{\sqrt{d}}
+ M_{\mathrm{causal}}
\right)V^{k}.
\end{equation}

\paragraph{Dual-Attention Block.}
Each DAE block integrates the two attention mechanisms sequentially with residual connections and Layer Normalization. 
Let $Z$ denote the input tensor to the block. The update process is formulated as:
\begin{align}
Z' &= Z + \text{IntraAttn}(Z) \label{eq:intra_update} \\
Z'' &= Z' + \text{InterAttn}(Z') \label{eq:inter_update} \\
Z_{\text{out}} &= Z'' + \text{MLP}(\text{LN}(Z'')), \label{eq:mlp_update}
\end{align}
where Eq.~\eqref{eq:intra_update} applies Intra-Frame Attention for each frame, and Eq.~\eqref{eq:inter_update} applies Inter-Frame Attention for each token sequence across time.
This sequential design ensures that the model captures both spatial consistency and temporal dynamics effectively.

\paragraph{Discussion.}
Intra-Frame Attention captures spatial relationships among Atomic Actions within each frame, while Inter-Frame Attention captures the temporal relationships of each local motion trajectory.
Together, these two attention mechanisms enable the DAE to learn transferable local motion representations.

\section{Experimental Details}
\label{appendix:implementation_details}

\subsection{Pre-training Dataset}
Following IPV, we adopt the Something-Something-V2 (SSV2) dataset for unsupervised pre-training. 
SSV2 contains more than 220K videos of humans interacting with everyday objects, covering a wide range of motion patterns and object manipulations. 
After filtering out videos with fewer than 25 frames, we retain 162K videos for pre-training. 
This large-scale and diverse dataset provides a rich data foundation for learning transferable local motion representations.

\subsection{Benchmark Environments}

\subsubsection{DMControl Remastered}
\begin{figure}
    \centering
    \includegraphics[width=0.50\textwidth]{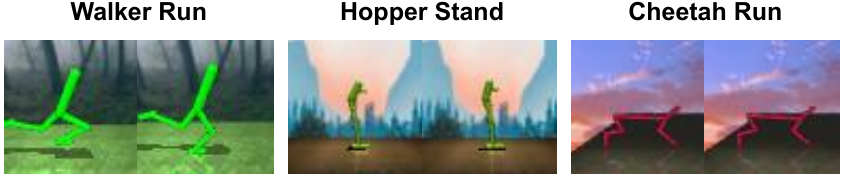}
    \caption{\textbf{Examples of DMControl Remastered.}}
    \label{fig:DMCR_vis}
\end{figure}

DMC Remastered (DMCR) is a variant of the DeepMind Control Suite featuring randomly generated graphics, emphasizing visual diversity.
In each episode, seven factors affecting visual conditions are randomly sampled, including background, floor texture, robot body color, target color, reflectance, camera position, and lighting.
Following IPV, we evaluate the same three locomotion tasks: ``Walker Run'', ``Hopper Stand'', and ``Cheetah Run'', as shown in Figure~\ref{fig:DMCR_vis}. 
Each episode lasts 1000 steps with an action repeat of~2, and rewards range from 0 to 1. 
All methods are trained for 1.02M environment steps to ensure fair comparison.

\begin{figure}
    \centering
    \includegraphics[width=0.50\textwidth]{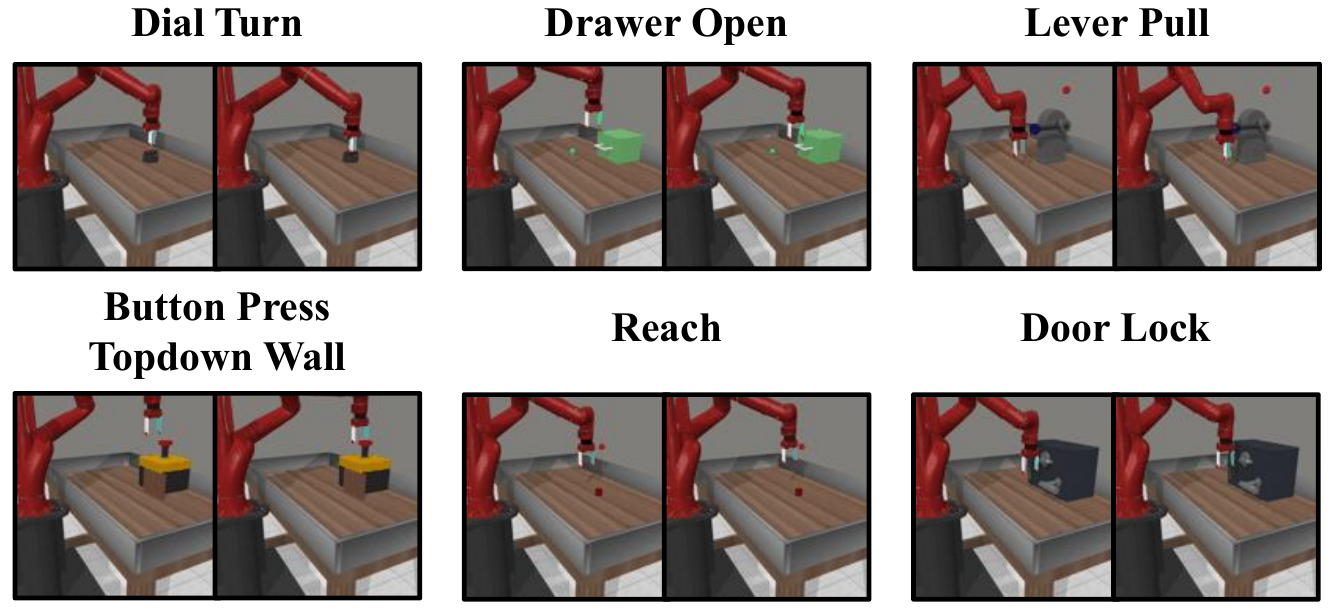}
    \caption{\textbf{Examples of Meta-World.}}
    \label{fig:metaworld_vis}
\end{figure}

\subsubsection{Meta-World}
Meta-World consists of 50 diverse robotic manipulation tasks. 
Following IPV and PreLAR, we evaluate six same tasks: ``Dial Turn'', ``Drawer Open'', ``Lever Pull'', ``Button Press Topdown Wall'', ``Reach'', and ``Door Lock'', as shown in Figure~\ref{fig:metaworld_vis}.
Each episode contains 500 steps with no action repetition, and the action space is 4-dimensional. 
Rewards range from 0 to 10. 
All methods are trained for 255K environment steps.

\begin{figure*}
    \centering
    \includegraphics[width=\textwidth]{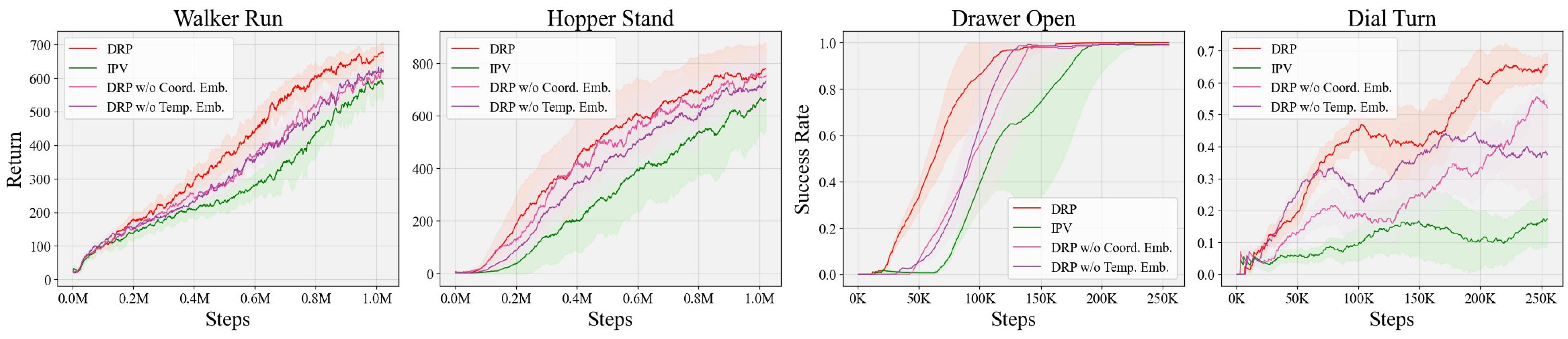}
    \caption{\textbf{Ablation study on Positional Embeddings in DAE.} We analyze the necessity of the two positional encodings within the DAE module. The results compare our full model DRP against variants where either the Coordinate Embedding (w/o Coord. Emb.) or the Temporal Embedding (w/o Temp. Emb.) is removed on the ``Walker Run'' and ``Hopper Stand'' tasks from DMCR, and the ``Drawer Open'' and ``Dial Turn'' tasks from Meta-World.}
    \label{fig:pos_emb_ablation_study}
\end{figure*}

\begin{figure}
    \centering
    \includegraphics[width=0.5\textwidth]{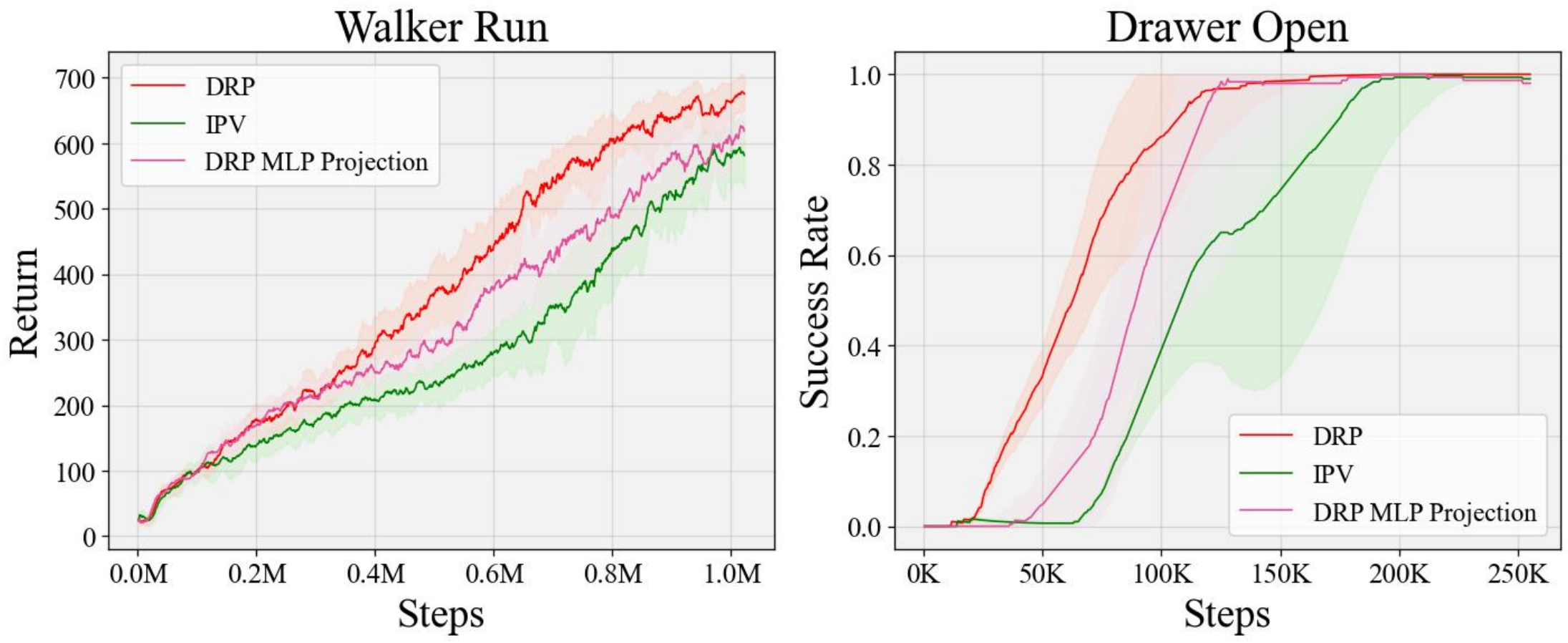}
    \caption{\textbf{Ablation study on the Adapter design.} We compare DRP with a variant where the Adapter is replaced by a simple Multi-Layer Perceptron (MLP) projection layer (denoted as ``DRP MLP Projection'') on the ``Walker Run'' task from DMCR and the ``Drawer Open'' task from Meta-World.}
    \label{fig:ablation_adapter}
\end{figure}

\section{Additional Ablation Studies}
\label{appendix:ablation}

We further evaluate the key design choices of DRP, including the two forms of positional embeddings (Coordinate Embedding and Temporal Embedding) used in the Dual-Attention Encoder (DAE) and the Adapter architecture.




\subsection{Ablation on Positional Embeddings in DAE}
\label{appendix:dae_pe}

When encoding Atomic Actions, the DAE augments the content embedding with two forms of positional information: 
(1) a \textit{coordinate embedding} that specifies where the flow patch is located, and 
(2) a \textit{temporal embedding} that indicates the timestep at which it occurs. 
These two positional cues are critical for enabling the DAE to capture the spatial and temporal relationships among Atomic Actions.

To evaluate the importance of these two positional encodings, we individually remove one type of positional encoding from the DRP.
As shown in Figure~\ref{fig:pos_emb_ablation_study}, ``DRP w/o Coord. Emb.'' denotes the variant without the Coordinate Embedding, and ``DRP w/o Temp. Emb.'' denotes the variant without the Temporal Embedding.
The experimental results indicate that removing either positional embedding leads to significant performance degradation.
This suggests that spatial position and temporal order information are critical for modeling the spatiotemporal relationships among atomic actions, and are therefore essential for learning transferable local motion representations.

\subsection{Ablation on the Adapter Design}
\label{appendix:adapter_ablation}
During the fine-tuning stage, the Adapter serves as a key bridge connecting the pre-trained latent dynamics model and the Action-Specific Dynamics Model.
To evaluate the necessity of the Adapter, we replace it with a simple Multi-Layer Perceptron (MLP) projection layer.

As shown in the Figure~\ref{fig:ablation_adapter}, the variant where the Adapter is replaced by an MLP projection layer, denoted as ``DRP MLP Projection,'' results in a significant performance drop.
This indicates that the MLP's limited expressive capacity prevents it from effectively mapping the local motion representations to the agent-specific action space.
This result highlights the critical role of the Adapter in effectively aligning the transferable local motion representation space with the agent-specific action space.
\section{Additional Visualization and Analysis}
\label{appendix:visualization}

In this section, we provide additional qualitative analyses to further validate the effectiveness of our proposed local motion representations.  
We present visualizations spanning three aspects: source-domain video prediction, image reconstruction, and t-SNE analysis of local motion representations.

\subsection{Source-Domain Video Prediction}
\label{appendix:source_pred_vis}




In addition to Figure~\ref{fig:video_prediction}(a) in the main manuscript, we provide more open-loop video prediction results on the source domain SSV2 test set to further evaluate the effectiveness of the learned local motion representations. 
Given initial video frames, the latent dynamics model performs open-loop predictions conditioned on the local motion representations.

\begin{figure*}
    \centering
    \includegraphics[width=\textwidth]{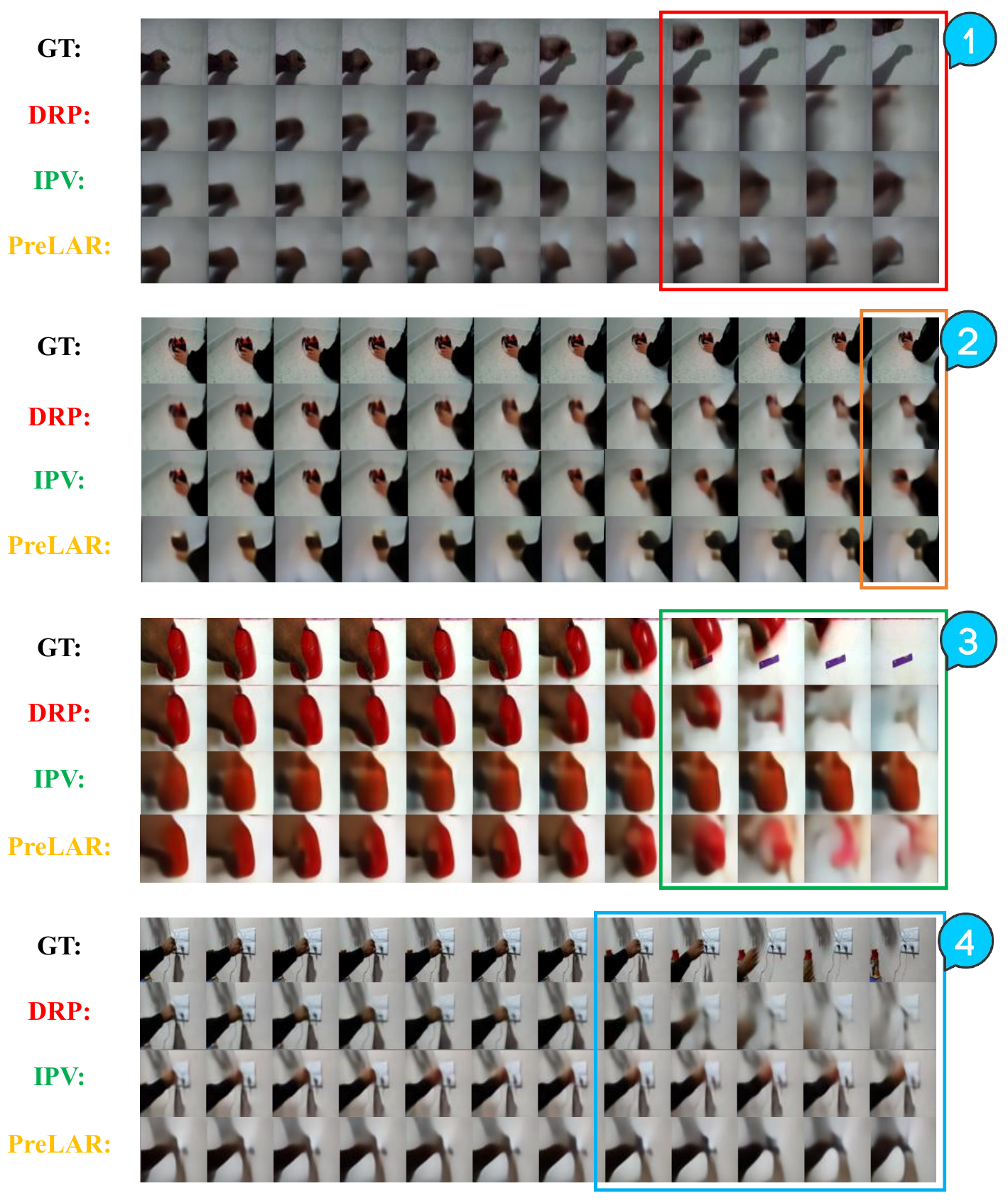}
    \caption{\textbf{Open-loop video prediction on the source domain (SSV2) test set.} We compare the prediction results of our method (DRP) with IPV and PreLAR, where ``GT'' denotes Ground Truth.}
    \label{fig:video_predict_supple}
\end{figure*}

As shown in Figure~\ref{fig:video_predict_supple}, our model generates more accurate future frame predictions compared to the baselines. Specifically:
(1) \textbf{Hand Motion:} DRP accurately captures the hand's movement and its subsequent exit from the frame, whereas the predictions from IPV and PreLAR fail to exhibit this motion.
(2) \textbf{Relative Positional Relationships:} In the process of pushing an object toward a wall corner, DRP accurately predicts the relative positional relationship as the object approaches the wall corner. In contrast, the subsequent predictions of IPV and PreLAR fail to exhibit this dynamic change in relative position. Furthermore, the wall corner disappears in IPV's later predictions, while PreLAR fails to capture the corner throughout the entire sequence.
(3) \textbf{Object Removal:} DRP accurately predicts the process of the hand and cup being removed from the view. Conversely, IPV and PreLAR incorrectly predict the cup as still visible. (Note: Since the object under the cup is occluded in the initial frames, DRP is unable to predict this purple object after the cup is removed, which is a reasonable and expected behavior).
(4) \textbf{Complex Manipulation:} DRP successfully captures the complex forward and backward motion in the plugging task. 
Both IPV and PreLAR fail to predict the backward movement of the arm.
Furthermore, arm deformation is observed in PreLAR's predictions.

These extended results further confirm that our learned local motion representations effectively capture robust and transferable motion patterns.




\subsection{Image Reconstruction Analysis}
\label{appendix:image_recon_vis}

\begin{figure*}
    \centering
    \includegraphics[width=\textwidth]{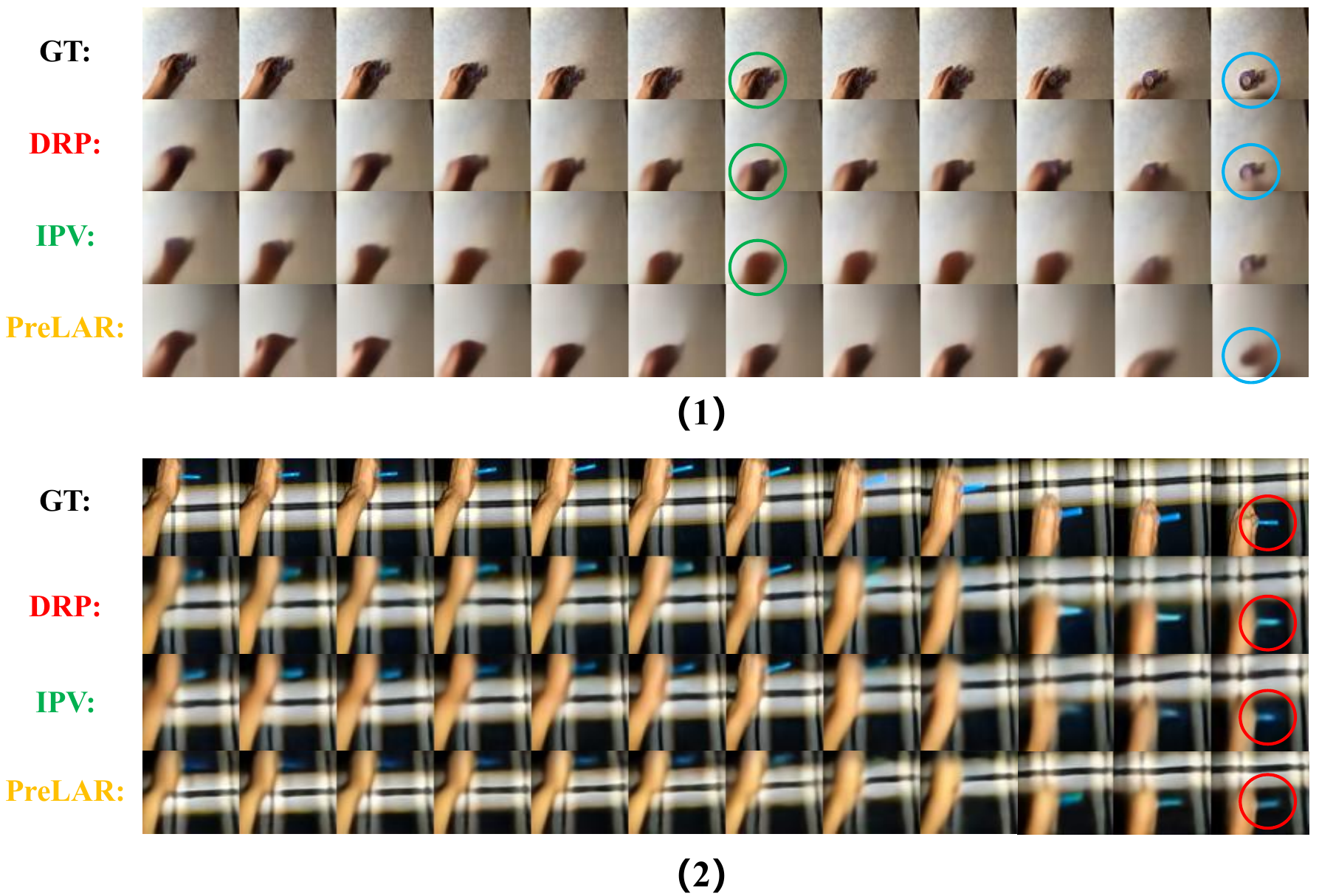}
    \caption{\textbf{Image reconstruction visualization from the latent dynamics model.} We compare the reconstruction results of our method (DRP) with IPV and PreLAR, where ``GT'' denotes Ground Truth.}
    \label{fig:image_recon_supple}
\end{figure*}

To further investigate the advantages of our proposed local motion representations, we visualize the image reconstruction results generated by the pre-trained latent dynamics model. 
As shown in Figure ~\ref{fig:image_recon_supple}, compared to the baseline model, our model's latent state captures finer visual details. 
Specifically:
\textbf{(1) Visual accuracy:} DRP accurately preserves the specific shape of the hand, whereas IPV generates a blurry reconstruction of hands. Similarly, DRP clearly captures the shape of the manipulated object, while PreLAR produces a blurry appearance.
\textbf{(2) Geometric accuracy:} DRP correctly maintains the object’s length, whereas IPV shortens it in the reconstruction. Moreover, DRP successfully captures the object’s orientation, whereas PreLAR fails to reproduce the correct orientation.

These results demonstrate that the latent dynamics model based on local motion representations is better at capturing motion-relevant visual features.

\begin{figure}
    \centering
    \includegraphics[width=0.5\textwidth]{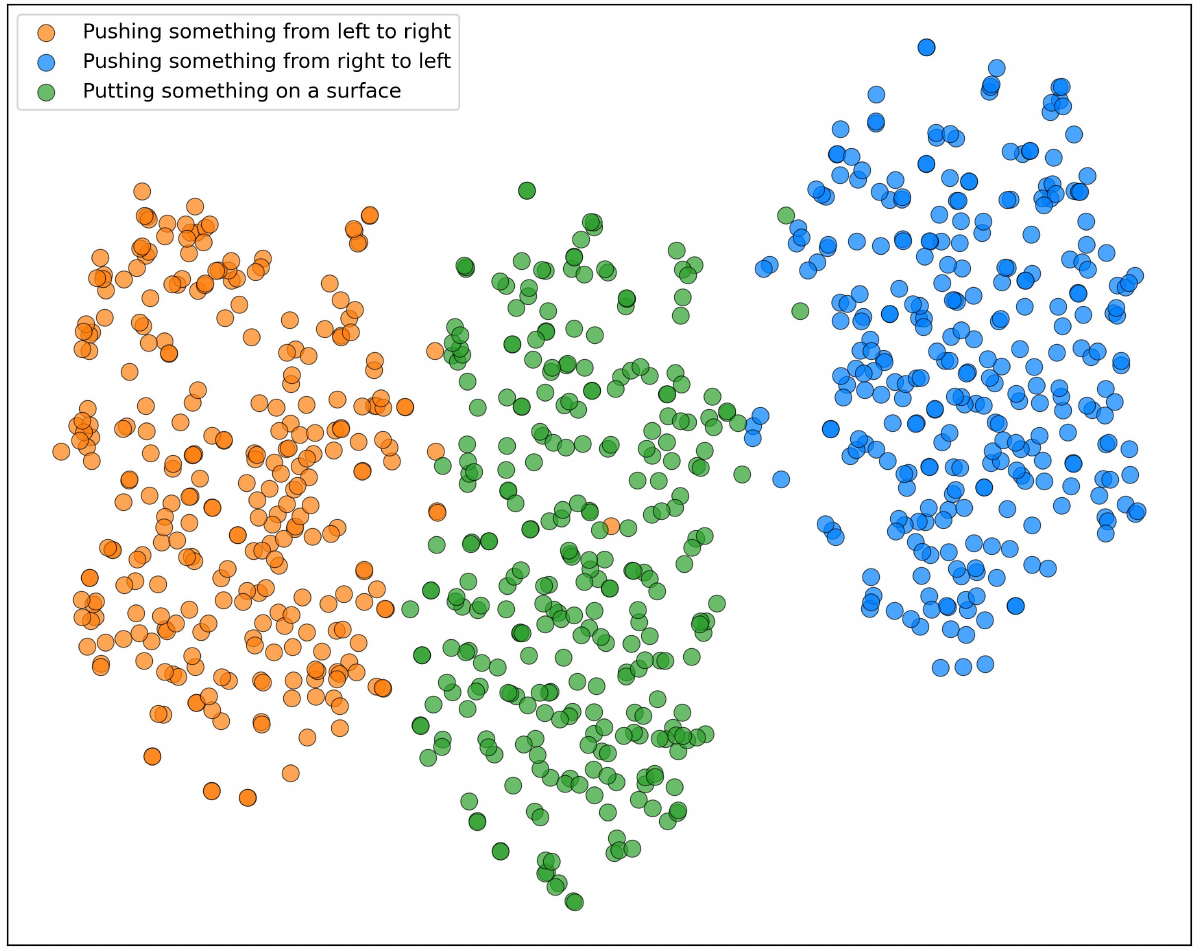}
    \caption{\textbf{t-SNE Visualization of Composed Local Motion Representations.} We perform t-SNE visualization on the composed local motion representations corresponding to three manipulation behaviors that share overall visual similarity (``Pushing something from left to right'', ``Pushing something from right to left'', and ``Putting something on a surface'').}
    \label{fig:t-SNE_vis_supple}
\end{figure}

\subsection{t-SNE Analysis of Local Motion Representations}
\label{ssec:tsne_analysis}

To further investigate the effectiveness of our proposed local motion representations, we perform t-SNE visualization on the recomposed local motion representations during pre-training, as shown in Figure~\ref{fig:t-SNE_vis_supple}.

Specifically, we sample video clips of length $T=25$ from the SSV2 dataset, which are respectively annotated with three different labels: ``Pushing something from left to right'', ``Pushing something from right to left'', and ``Putting something on a surface''. 
We then visualize the recomposed local motion representations of these clips using t-SNE, as shown in Figure~\ref{fig:t-SNE_vis_supple}.
Notably, no video labels are utilized during the pre-training stage.
Although these three manipulation behaviors exhibit overall visual similarity (especially the two pushing actions in opposite directions), their corresponding recomposed local motion representations form clearly distinct clusters in the latent space.

This visualization result strongly demonstrates that our learned local motion representations are capable of effectively capturing and distinguishing the agent's different motion patterns.





\section{Hyperparameter Analysis}
\label{appendix:hyper}
Apart from the newly introduced hyperparameters: the number of tracked keypoints $K$, the MAE mask ratio $\rho$, the local optical flow patch size $P$, and the number of Dual-Attention Blocks $L$, we adopt the same hyperparameter settings as IPV and PreLAR to ensure a fair comparison. 
The complete set of hyperparameters is provided in Table~\ref{table:hyperparameters}.

\subsection{Number of Tracked Motion Keypoints}
\label{appendix:kstudy}

This section analyzes the selection of the hyperparameter $K$, the number of motion keypoints tracked in our Atomic Action Extraction module.
We investigate the influence of $K$ by varying its value across $K \in \{8, 16, 32\}$. 
These hyperparameter experiments are conducted on the DMCR tasks ``Walker Run'' and ``Hopper Stand'', with the results summarized in Table~\ref{table:ablation_keypoints_num}.

As shown in the results, selecting an insufficient number of keypoints ($K=8$) leads to suboptimal performance. 
This is likely because a small $K$ fails to adequately capture all meaningful local motion components that constitute the global motion.
Conversely, increasing the number of tracked keypoints from $K=16$ to $K=32$ yields minimal performance improvement, and even results in a slight decrease on the ``Walker Run'' task. 
Crucially, this increase significantly improves the overall computational cost.

Therefore, we select $K=16$ as the default number of tracked motion keypoints in this work, as it strikes the balance between performance and computational cost.


\begin{table}
\setlength\tabcolsep{2pt}
\caption{\textbf{Hyperparameter analysis on the Keypoints number $K$.}}
\centering
\resizebox{0.4\textwidth}{!}{%
\begin{tabular}{l | c c c}
    \toprule
    Keypoints number $K$ & $K=8$ & $K=16$ & $K=32$ \\
    \midrule
     Walker Run & 646 $\pm$ 45 & \textbf{681 $\pm$ 39} & 667$\pm$ 43 \\
    \midrule
     Hopper Stand & 775 $\pm$ 123 & 796 $\pm$ 114 & \textbf{799$\pm$ 105} \\
    \bottomrule
\end{tabular}%
}
\label{table:ablation_keypoints_num}
\end{table}

\begin{table}
\setlength\tabcolsep{2pt}
\caption{\textbf{Hyperparameter analysis on the MAE mask ratio $\rho$.}}
\centering
\resizebox{0.4\textwidth}{!}{%
\begin{tabular}{l | c c c}
    \toprule
    Mak Ratio $\rho$ & $\rho=0.3$ & $\rho=0.5$ & $\rho=0.7$ \\
    \midrule
     Walker Run & 648 $\pm$ 42 & \textbf{681 $\pm$ 39} & 660$\pm$ 47 \\
    \bottomrule
\end{tabular}%
}
\label{table:ablation_mask_ratio}
\end{table}

\subsection{MAE Mask Ratio}
\label{appendix:maskratio}

We investigate the influence of the MAE masking ratio $\rho$ by evaluating $\rho \in \{0.3, 0.5, 0.7\}$.
These hyperparameter experiments are conducted on the DMCR tasks ``Walker Run'', with the results summarized in Table~\ref{table:ablation_mask_ratio}.

A low masking ratio ($\rho=0.3$) reduces the reconstruction difficulty, which may fail to learn robust spatiotemporal relationships. 
Conversely, an excessively high masking ratio ($\rho=0.7$) leads to excessive information loss, making effective reconstruction impossible.
As shown in Table~\ref{table:ablation_mask_ratio}, a ratio of $0.5$ achieves the optimal performance.

\begin{table}
\setlength\tabcolsep{2pt}
\caption{\textbf{Hyperparameter analysis on the local optical flow patch size $P$.}}
\centering
\resizebox{0.4\textwidth}{!}{%
\begin{tabular}{l | c c c}
    \toprule
    Patch Size $P$ & $P=12$ & $P=16$ & $P=24$ \\
    \midrule
     Walker Run & 649 $\pm$ 52 & \textbf{681 $\pm$ 39} & 652$\pm$ 32 \\
    \bottomrule
\end{tabular}%
}
\label{table:ablation_patch_size}
\end{table}

\subsection{Local Patch Size}
\label{Hyperparameter:mask_ratio}

We evaluate the influence of the local optical flow patch size $P$ by varying its value across $P \in \{12, 16, 24\}$.
These hyperparameter experiments are conducted on the DMCR task ``Walker Run'', with the results summarized in Table~\ref{table:ablation_patch_size}.

A smaller patch size ($P=12$) results in an excessively localized receptive field.
Such localized features tend to exhibit high variance when capturing motion patterns and are highly sensitive to small fluctuations, leading to representations that lack robustness.
Conversely, an excessively large patch size ($P=24$) implies insufficient deconstruction, where a single patch may include multiple distinct local motion patterns, thereby hindering effective cross-domain transfer.
The experimental results demonstrate that $P=16$ balances precise local motion capture and representation robustness.

\begin{table*}
\centering
\caption{\textbf{Hyperparameters in our experiments.} We use the same hyperparameter setting as IPV.}
\begin{tabular}{l c c c}
    \toprule
     & Hyperparameter & Value\\
    \midrule
    \multirow{9}{*}{Pre-training
from Videos} & Image size & $64 \times 64 \times 3$ \\
    & Image preprocess & Linearly rescale from {[}0, 255{]} to {[}-0.5, 0.5{]} \\
    & Video segment length $T$ & 25 \\
    & KL weight $\beta_z$ & 1.0 \\
    & Optimizer & Adam \\
    & Learning rate & 3 $\times$ 10$^{-4}$ \\
    & Batch size & 16 \\
    & Training iterations & 6 $\times$ 10$^5$ \\
    & Number of tracked keypoints $K$ & 16 \\
    & Local optical flow patch size $P$ & 16 \\
    & MAE mask ratio $\rho$ & 0.5 \\
    & Number of Dual Attention Blocks $L$ & 6 \\
    \midrule
    \multirow{27}{*}{Fine-tuning
with MBRL} & Observation size & $64 \times 64 \times 3$ \\
    & Observation preprocess & Linearly rescale from {[}0, 255{]} to {[}-0.5, 0.5{]} \\
    & Trajectory segment length $T$ & 50 \\
    & \multirow{2}{*}{Random exploration} & 5000 environment steps for Meta-world \\
    & & 1000 environment steps for DMCR \\
    & Replay buffer capacity & 10$^6$ \\
    & Training frequency & Every 5 environment steps \\
    & Action-conditional KL weight $\beta_s$ & 1.0 \\
    & Representative reward predictor weight $\beta_r$ & 1.0 \\
    & \multirow{2}{*}{Intrinsic reward weight $\lambda$} & 1.0 for Meta-World \\
    & & 0.1 for DMCR  \\
    & Imagination horizon $H$ & 15 \\
    & Discount $\gamma$ & 0.99 \\
    & $\lambda$-target discount & 0.95 \\
    & Entropy regularization $\eta$ & 1 $\times$ 10$^{-4}$ \\
    & \multirow{2}{*}{Batch size} & 50 for Meta-World \\
    & & 16 for DMCR \\
    & World model optimizer & Adam \\
    & World model learning rate & 3 $\times$ 10$^{-4}$ \\
    & Actor optimizer & Adam \\
    & Actor learning rate & 8 $\times$ 10$^{-5}$ \\
    & Critic optimizer & Adam \\
    & Critic learning rate & 8 $\times$ 10$^{-5}$ \\
    & Evaluation episodes & 10 \\
    & Number of tracked keypoints $K$ & 16 \\
    & Local optical flow patch size $P$ & 16 \\
    & MAE mask ratio $\rho$ & 0.5 \\
    & Number of Dual Attention Blocks $L$ & 6 \\
    \bottomrule
\end{tabular}

\label{table:hyperparameters}
\end{table*}


\end{document}